\documentclass[final,5p,times,twocolumn]{elsarticle}

\usepackage{amssymb}
\usepackage{amsmath}
\usepackage{amsthm}
\usepackage{multirow}
\journal{Neurocomputing}
\usepackage{lineno}
\usepackage{booktabs}
\usepackage{subcaption}
\usepackage{tabularx}

\begin{document}

\begin{frontmatter}

\title{SemCo: Toward Semantic Coherent Visual Relationship Forecasting}

\author[label1]{Yangjun Ou\fnref{cofirst}}
\ead{yjou@wtu.edu.cn}

\author[label1]{Yao Liu\fnref{cofirst}}
\ead{2415283050@wtu.edu.cn}

\author[label2]{Li Mi\corref{cor1}}
\ead{li.mi@ethz.ch}

\fntext[cofirst]{Equal Contribution}
\cortext[cor1]{Corresponding Author}

\author[label3]{Zhenzhong Chen}
\ead{zzchen@ieee.org}

\affiliation[label1]{
            organization={School of Computer Science and Artificial Intelligence},
            addressline={Wuhan Textile University}, 
            city={Wuhan},
            country={China}
            }
\affiliation[label2]{
            organization={ETH Zürich},
            city={Zürich},
            country={Switzerland}
            }
\affiliation[label3]{
            organization={School of Remote Sensing and Information Engineering},
            addressline={Wuhan University}, 
            city={Wuhan},
            country={China}
            }

\begin{abstract}
Visual Relationship Forecasting (VRF) aims to anticipate relations among objects without observing future visual content. 
The task relies on capturing and modeling the semantic coherence in object interactions, as it underpins the evolution of events and scenes in videos.
However, existing VRF datasets offer limited support for learning such coherence due to noisy annotations in the datasets and weak correlations between different actions and relationship transitions in subject-object pair. Furthermore, existing methods struggle to distinguish similar relationships and overfit to unchanging relationships in consecutive frames.
To address these challenges, we present SemCoBench, a benchmark that emphasizes semantic coherence for visual relationship
forecasting. Based on action labels and short-term subject-object pairs, SemCoBench decomposes relationship categories and dynamics by cleaning and reorganizing video datasets to ensure predicting semantic coherence in object interactions. In addition, we also present Semantic Coherent Transformer method (SemCoFormer) to model the semantic coherence with a Relationship Augmented Module (RAM) and a Coherence Reasoning Module (CRM). RAM is designed to distinguish similar relationships, and CRM facilitates the model's focus on the dynamics in relationships.
The experimental results on SemCoBench demonstrate that modeling the semantic coherence is a key step toward reasonable, fine-grained, and diverse visual relationship forecasting, contributing to a more comprehensive understanding of video scenes.
\end{abstract}

\begin{keyword}
Visual Relationship Forecasting \sep Future Prediction \sep Visual Reasoning
\end{keyword}

\end{frontmatter}

\section{INTRODUCTION}
\label{sec:introduction}
Future prediction aims at generating future video frames~\cite{su2017predicting, lee2017desire, luc2017predicting}, predicting object trajectories~\cite{pellegrini2009you,lerner2007crowds,robicquet2016learning}, or anticipating future actions~\cite{yeung2018every,damen2018scaling}, based only on the past observations. By capturing the visual or semantic coherence in videos, future prediction tasks reason about the future scenes or events, which has become an important research topic in video understanding and analysis~\cite{zhou2018end, Zhao2020Open}.

\begin{figure}[t]
	\centering
	\includegraphics[width=0.48\textwidth]{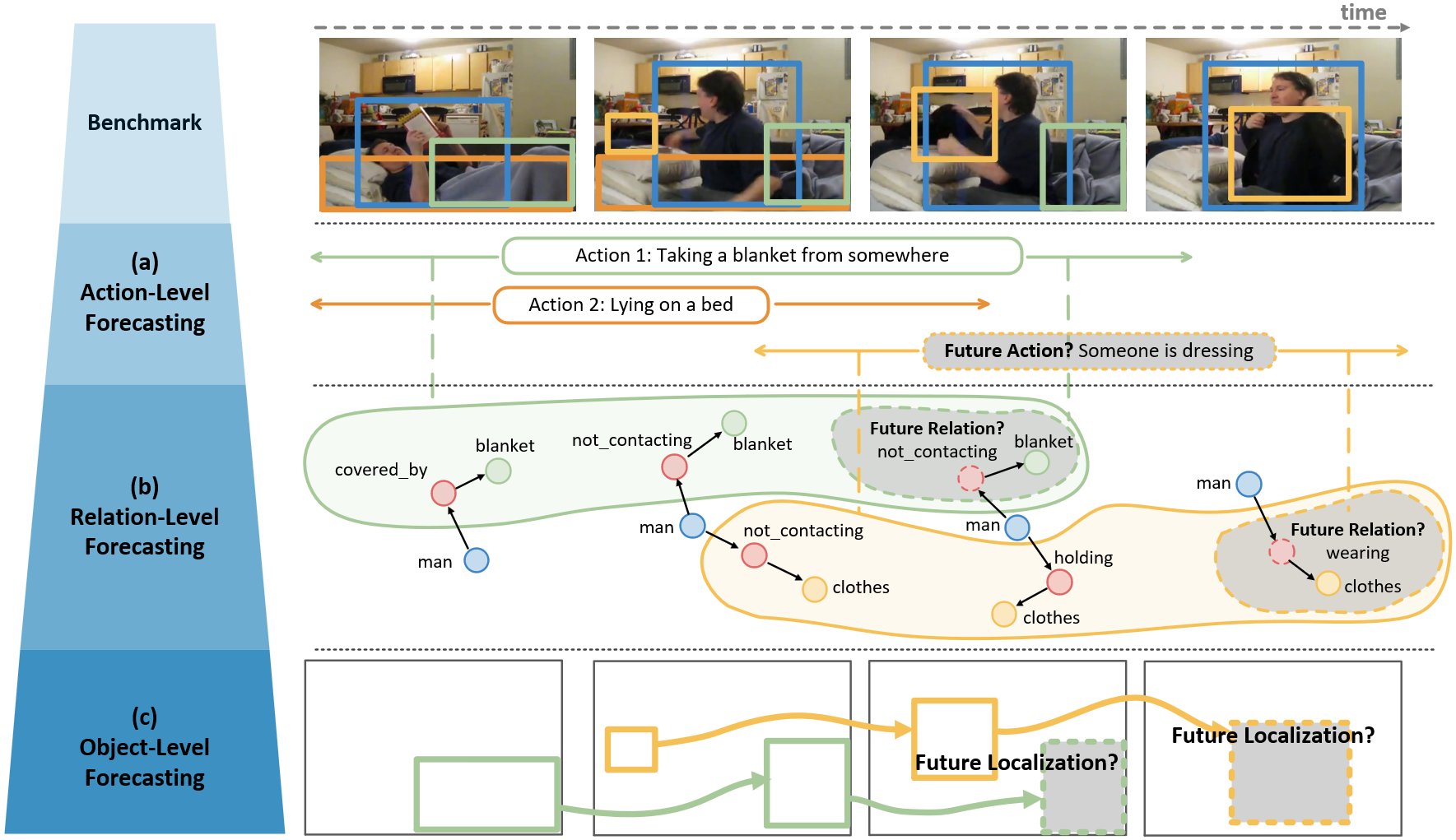}
	\caption{The illustration of the action-level forecasting tasks (a), object-level forecasting task (c) and the relation-level forecasting task (b). Different from the action-level and object-level forecasting, relation-level forecasting is conducted to predict future interactions between objects on a fine-grained time scale based on the semantic coherence to achieve an action or event.}
	\label{fig:example}
\end{figure}

Future prediction tasks (as shown in Figure~\ref{fig:example}) can be categorized into different levels of granularity. One line of work focuses on generating future video frames or predicting object trajectories based on visual coherence~\cite{su2017predicting, robicquet2016learning} (Figure~\ref{fig:example} (a)). Some other tasks go beyond visual coherence and emphasize semantic coherence, aiming to forecast future actions or object states~\cite{yeung2018every, damen2018scaling} (Figure~\ref{fig:example} (c)), where semantic coherence concerns the plausibility of action evolution. However, action sequences or video events are often weakly correlated over time, making them challenging and less predictable.
For instance, in Figure~\ref{fig:example}, predicting the future action `\textit{Someone is dressing}' from the two prior actions is hard because the related object \textit{clothes} is nearly invisible in earlier frames. \par
\begin{figure}[t]
	\centering
	\includegraphics[width=0.48\textwidth]{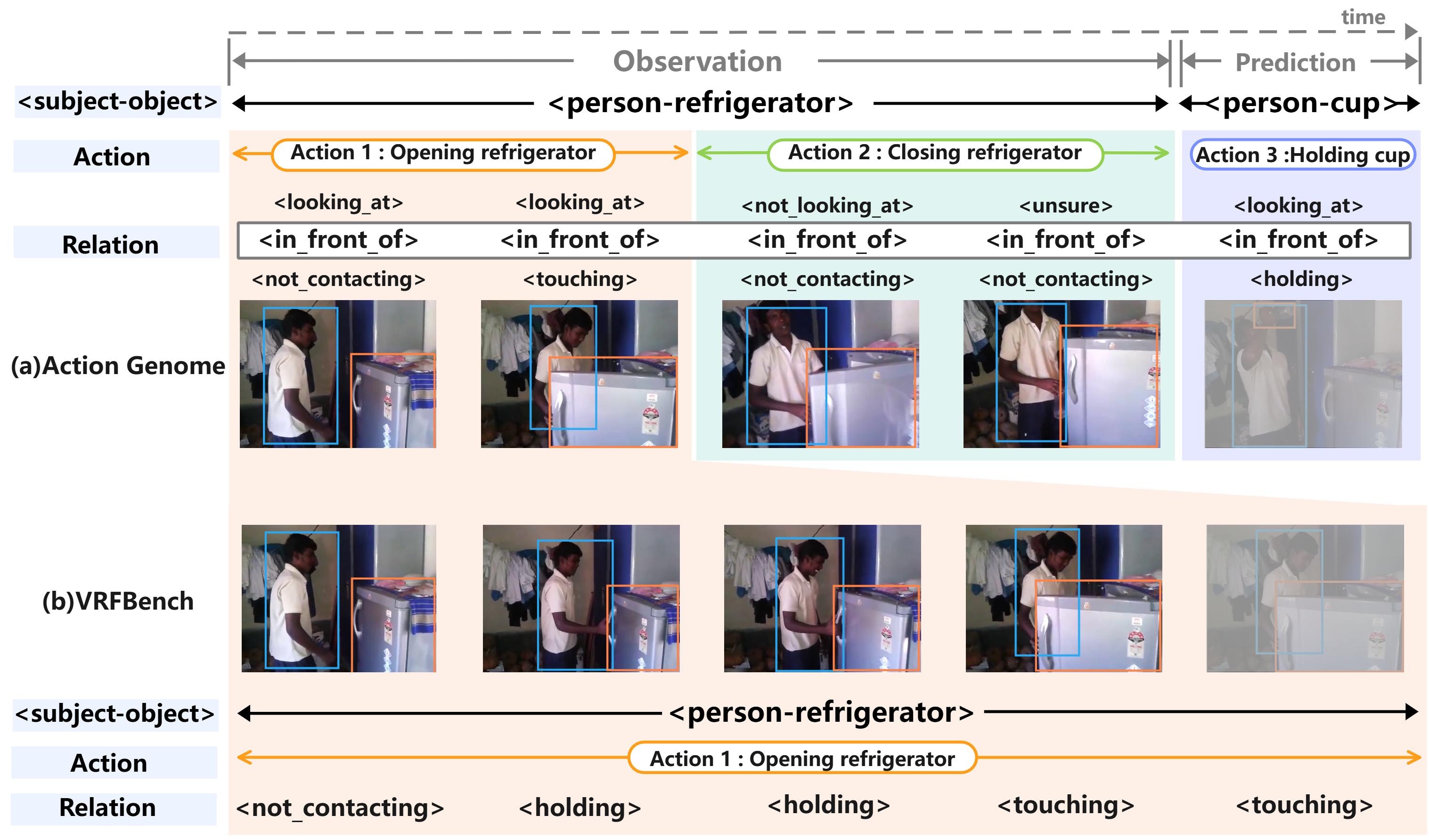}
	\caption{\textbf{Comparison of the (a) previous dataset (\textit{i.e.}, Action Genome dataset) and (b) the proposed dataset for VRF task.} (a) The Action Genome dataset requires predicting relations that are associated with different action labels (\textit{e.g.}, `\textit{Opening refrigerator}', `\textit{Closing refrigerator}', and `\textit{Holding}') or subject-object pairs (\textit{e.g.}, person-refrigerator, person-cup) or relations that are not linked to certain actions (\textit{e.g.}, \textit{in front of}) (b) The proposed SemCoBench. Each sample is associated with an action (\textit{e.g.}, `\textit{Opening refrigerator}') or a subject-object pair (\textit{e.g.}, person-refrigerator) in a short period of video to ensure the semantic coherence of relation dynamics.}
	\label{fig:constract}
\end{figure}
Unlike actions and events, predicting relation-level forecasting research can provide semantic coherence in object interactions, ensuring reasonable relationship transitions~\cite{cao20213,qu2023dual, shang2017video, ji2020action} (Figure~\ref{fig:example} (b)). Specifically, actions or events in a short period can be represented as a sequence of semantically associated visual relationships, shown as $\textless$subject-predicate-object$\textgreater$. For example, an action `\textit{Someone is dressing}' can be decomposed as a series of visual relationships between person and clothes, such as $\textless$ \textit{person-holding-clothes} $\textgreater$ and $\textless$ \textit{person-wearing-clothes} $\textgreater$. Within the action, the relationship transitions are closely related and thus feasible to anticipate. 

Current relation-level forecasting research is mainly performed on video scene graph datasets (\textit{e.g.}, Action Genome~\cite{ji2020action}), which still fall short in capturing the semantic coherence between object interactions. On one side, relations annotated in those datasets are usually noisy and less predictable. 
For example, in Figure~\ref{fig:constract} (a), spatial relationships like \textit{in front of} do not change within the video clip and therefore do not indicate a certain action or video scene. On the other side, current efforts usually perform the forecasting task over a long period. During the video, subject-object pairs and actions might change (Figure~\ref{fig:constract} (a)), which may disturb the models' attention from reasoning the semantic coherences of the relations over time.

To address the challenges above, we introduce a new benchmark, named SemCoBench. The proposed SemCoBench consists of two datasets, SemCo-AG and SemCo-VidOR, which densely annotate 13 and 35 visual relationships in 1,923 and 13,447 video clips, respectively. Compared with current datasets for VRF, the proposed SemCoBench has two characteristics: 1) relation annotations are \textbf{less noisy and more predictable}: The relation categories that are visible in the video frames and associated with certain actions are selected manually. 2) relation transactions are \textbf{closely-related}: We anticipate future interactions between a defined subject-object pair within a certain action or a short period. Avoid focusing on interactions that are not relevant to semantic coherence. For example in Figure~\ref{fig:constract} (b), given a sequence of previous relationship triplets between \textit{person} and \textit{refrigerator}, our goal is to anticipate future triplets ($\textless$ \textit{person-touching-refrigerator} $\textgreater$) when the action of `\textit{Opening refrigerator}' took place.
\begin{table*}[t]
	\centering
	\caption{Comparison of the Public Future Forecasting Datasets.}
    \vspace{1mm}
    \resizebox{0.86\textwidth}{!}{
	\begin{tabular}{lcccc}
		\toprule
		Dataset & Task  & Samples & Annotations  & Scenes  \\
		\midrule
		
		ETH\cite{pellegrini2009you} & Trajectory Forecast & - & 750 trajectories & Ourdoor \\
		UCY\cite{lerner2007crowds} & Trajectory Forecast & - & 786 trajectories & Ourdoor \\
		Stanford Drone\cite{robicquet2016learning} & Trajectory Forecast & - & 10K trajectories & Ourdoor \\
		Human Interaction Videos\cite{fan2018forecasting} & Location Forecast & 47 videos & - & Indoor \\
		Activities of Daily Living\cite{pirsiavash2012detecting} & Location Forecast & 20 videos & - & Indoor \\
		Citywalks\cite{styles2020multiple} & Location Forecast & 358 videos & - & Ourdoor \\
		MultiTHUMOS\cite{yeung2018every} & Action Forecast  & 400 videos & 38.7K action segments & Outdoor \\
		Epic Kitchens\cite{damen2018scaling} & Action Forecast & 10K videos & 39.6K action segments & Indoor \\
        Action Genome\cite{ji2020action}& Scene Graph Anticipation & 10K videos & 1.7M visual relationships & Indoor \\
        
		\midrule
		SemCo-VidOR $\&$ SemCo-AG & Visual Relationship Forecast & 15K videos & 135.8K visual relationships & Indoor $\&$ Outdoor \\
		\bottomrule
	\end{tabular}%
    }
	\label{tab:2.0}%
\end{table*}%

In addition, existing methods excel at modelling the evolution of dynamic interactions between objects, yet fail to account for the semantic coherence of object interactions. On one hand, the interaction between visual cues and the semantic cues of the object itself can help models distinguish similar relationships, yet existing methods often overlook this aspect. For instance, relying solely on visual cues makes it difficult to differentiate between \textit{holding} and \textit{carrying}, but the semantic cue of $\textless$\textit{person-refrigerator}$\textgreater$ can identify the \textit{holding} relationship. 
On the other hand, these approaches overfit to invariant relationships across consecutive frames, causing predictions to remain stuck at superficially stable relationships rather than reflecting the semantic cohenrence of object interactions.

We present Semantic Coherent Transformer method (SemCoFormer) for semantic coherent visual relationship forecasting. We argue that inferring predicate relations requires the interaction of visual and semantic cues from subject-object pairs. For instance, integrating the visual and semantic cues of $\textless$\textit{person-refrigerator}$\textgreater$ helps models learn the $\textless$\textit{person-holding-refrigerator}$\textgreater$ relationship. Based on this, we propose a Relationship Augmented Module (RAM), which augments the visual and semantic features of subjects and objects and serves as the foundation for inferring semantic coherence in object interactions. Furthermore, recognizing that relationship transitions are more obvious in cross-frames than in consecutive ones, we propose a Coherence Reasoning Module (CRM). CRM employs a sparse coding strategy, which enables the model to capture relational dynamics across non-consecutive frames, thereby predicting future relationships that better correspond to the semantic coherence. Experimental results on both SemCo-AG and
SemCo-VidOR datasets demonstrate that SemCoFormer outperforms the state-of-the-art baseline methods on visual relationship forecasting.

The contribution of the paper can be summarized as:
\begin{itemize}
	\item {We propose SemCoBench, a benchmark designed for semantic coherent visual relationship forecasting. Two datasets with closely-related, spatio-temporally localized visual relation annotations are established for advancing studies in visual relationship forecasting and video understanding.}
	\item {A Semantic Coherent Transformer (SemCoFormer) is proposed to capture semantic coherence in object interactions and serves as a baseline method on the proposed SemCoBench. SemCoFormer leverages cross-modal cues to extract coherent features and applies sparse coding to facilitate the model's focus on cross-frame relationship dynamics. SemCoFormer outperforms the state-of-the-art VRF methods on forecasting semantic coherent visual relationship transitions.} 
\end{itemize}

\section{BACKGROUND AND PRELIMINARIES}
\label{sec:related_work}

\subsection{Future Prediction Task and Benchmarks}
Future prediction in video can be roughly divided into three themes: 1) \textit{generating future frames or trajectories} (\textit{e.g.}, trajectory forecasting~\cite{pellegrini2009you,lerner2007crowds,robicquet2016learning}, location forecasting, and video forecasting~\cite{voleti2022mcvd}); 2) \textit{predicting future labels or states} (\textit{e.g.}, future human behavior~\cite{cao20213}, action~\cite{gong2022future,mascaro2023intention} or object state forecasting~\cite{yeung2018every,damen2018scaling}) and 3) scene graph and visual relationship~\cite{peddi2024towards,peddi2025towards,nguyen2024hyperglm,ouyang2023multiple} anticipation. For the first theme, tasks were proposed to forecast object trajectories, object locations or video frames in the unseen future, e.g., trajectory forecasting~\cite{robicquet2016learning,alberico2025egocentric}, location forecasting, and video forecasting~\cite{voleti2022mcvd}. In the second theme, future actions or object states are predicted based on a series of past observations~\cite{yeung2018every,damen2018scaling}. The third category of research focuses on predicting the interaction relationships between subject-object pairs in future video frames based on video frames or image sequences~\cite{peddi2024towards,peddi2025towards,nguyen2024hyperglm}.

To achieve future prediction, several datasets are proposed for different tasks as shown in Table \ref{tab:2.0}. 
For the trajectory forecasting task, Stanford Drone dataset~\cite{robicquet2016learning} includes a series of videos captured by a hovering drone from 8 different college campuses. All trajectories are provided in 2D coordinates and include other objects like bikes or cars apart from pedestrians.
For the location forecasting task, the Citywalks dataset~\cite{styles2020multiple} is proposed to predict future bounding boxes of tracked objects. In contrast to existing works on the trajectory forecasting task primarily from a birds-eye perspective, the object location forecasting task formulates the problem from an object-level perspective and calls for the prediction of full object bounding boxes, rather than trajectories alone.
For the action forecasting task, the MultiTHUMOS dataset~\cite{yeung2018every} is extended from the THUMOS dataset for action prediction, which contains dense, multi-label, frame-level action annotations. The MultiTHUMOS dataset aims to make predictions about what is likely to happen next or what happened previously in the video.
The EPIC-KITCHENS dataset~\cite{damen2018scaling} is a large-scale egocentric video benchmark recorded by 32 participants in their native kitchen environments. EPIC-KITCHENS dataset aims to anticipate the actor’s next action from a first-person view.

Unlike action-level or object-level forecasting, relation-level forecasting serves as an intermediate stage to bridge action~\cite{mascaro2023intention} and object~\cite{sun2019relational} in video understanding. 
The visual relationship forecasting task belongs to relation-level forecasting. However, unlike scene graph anticipation, which predicts scene graph to represent video scene evolution, the visual relationship forecasting task focuses on understanding and predicting relation transitions between subject-object triplets. This finer-grained formulation allows for a more precise understanding of dynamic relationships and often yields better interpretability. Current researches in relation-level forecasting use video scene graph datasets, which may contain noisy and unreasonable relation labels for anticipation. In this paper, we propose a new benchmark for visual relationship forecasting, with manually selected relation categories and closely correlated relation dynamics for fine-grained video scene understanding.

\begin{figure*}[t]
	\centering
	\includegraphics[width=\textwidth]{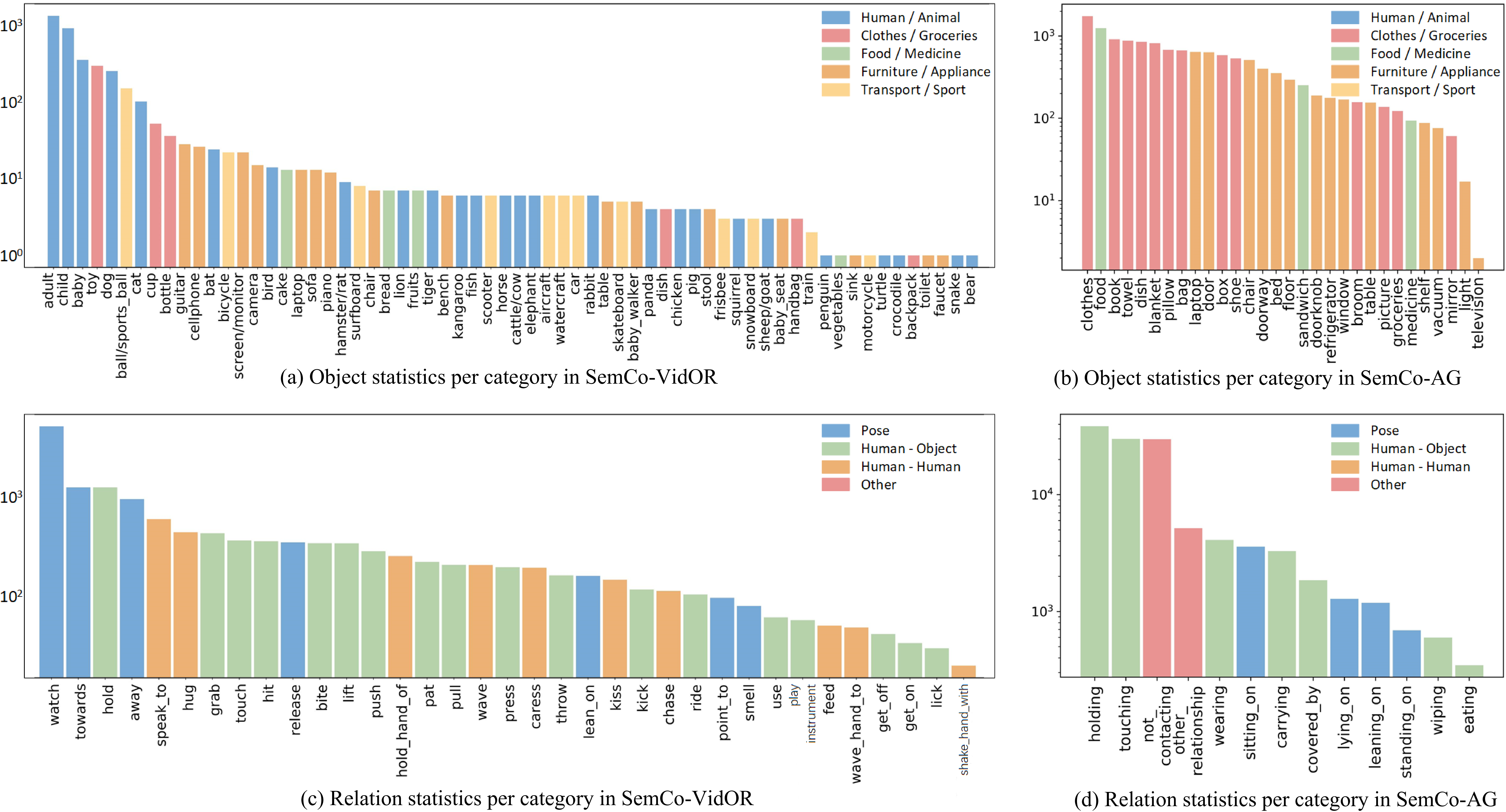}
	\caption{Object and relation statistics per category with colors indicating different types in SemCo-VidOR and SemCo-AG datasets.}
	\label{fig:dataset}
\end{figure*}
\begin{figure}[t]
	\begin{center}
		\includegraphics[width=0.48\textwidth]{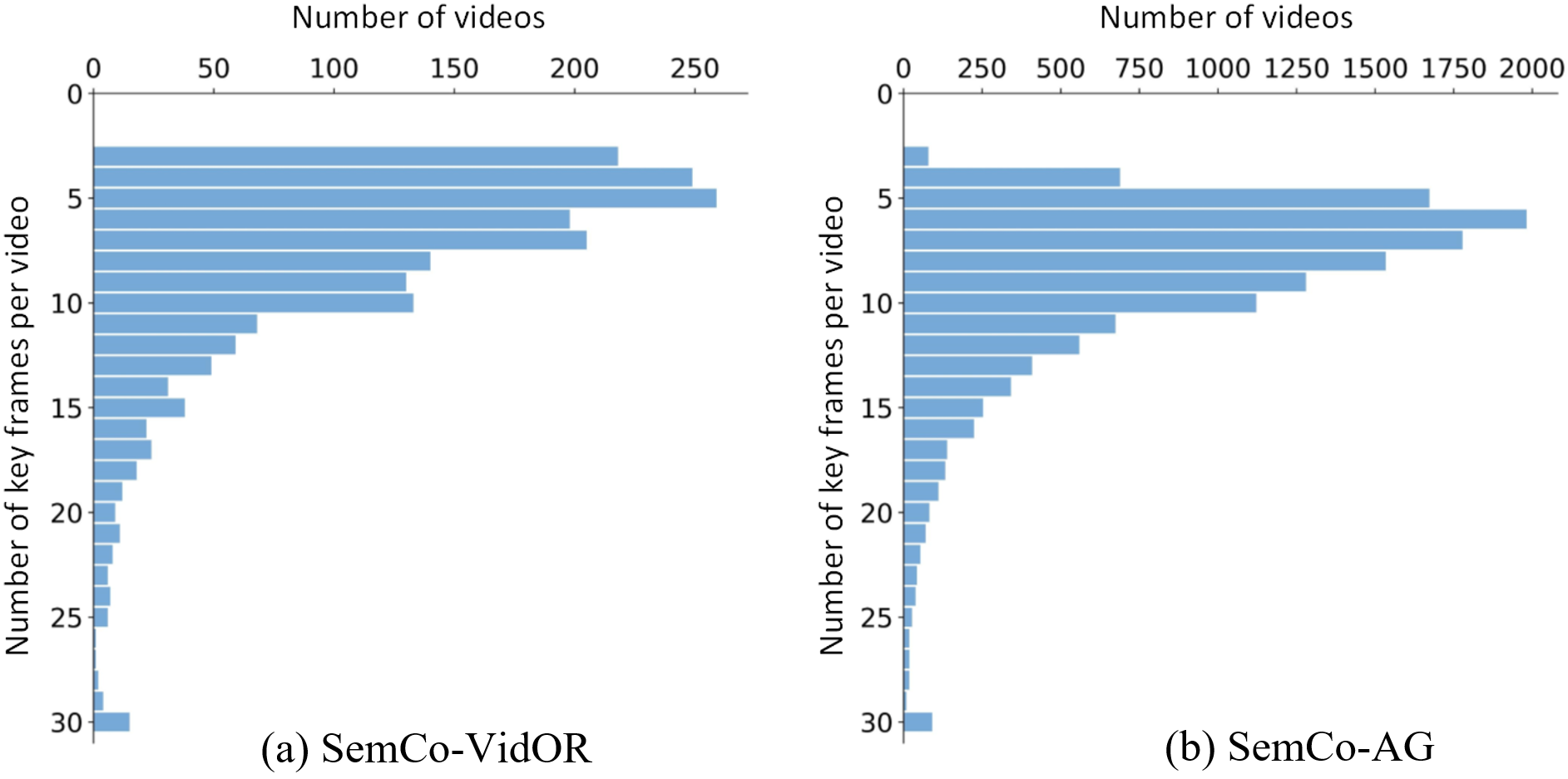}
	\end{center}
	\caption{{The number of keyframes per video in SemCo-VidOR and SemCo-AG datasets. The horizontal axis represents the number of videos, and the vertical axis represents the number of keyframes per video.}}
	\label{fig:datasettime}
\end{figure}
\begin{figure*}[t]
	\begin{center}
		\includegraphics[width=0.88\textwidth]{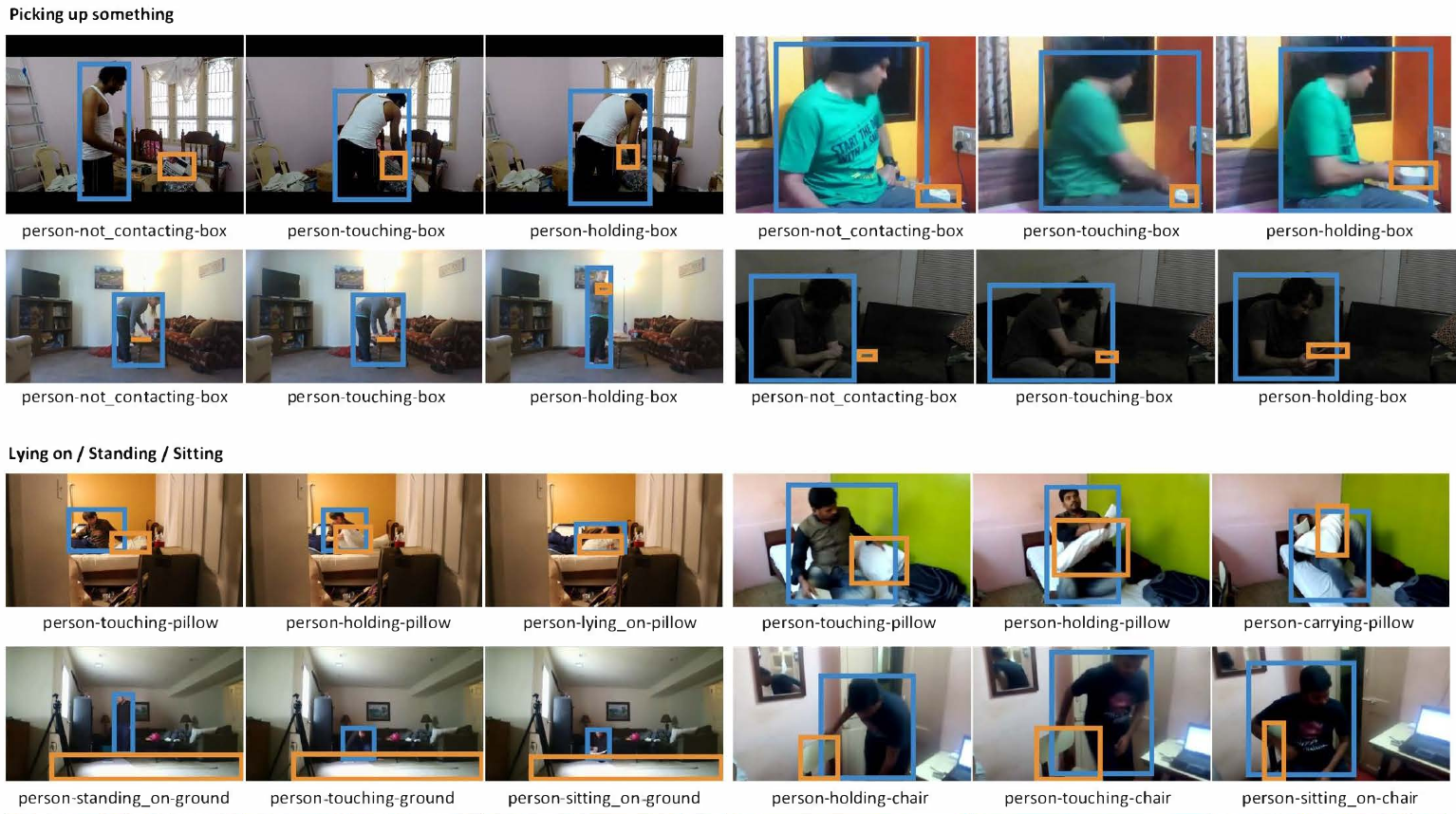}
	\end{center}
	\caption{Examples of the SemCo-AG datasets. Selected video frames of \textit{Picking up something} and \textit{Lying, Standing, Sitting} are showed.}
	\label{fig:example1}
\end{figure*}
\begin{figure*}[t]
	\begin{center}
		\includegraphics[width=0.88\textwidth]{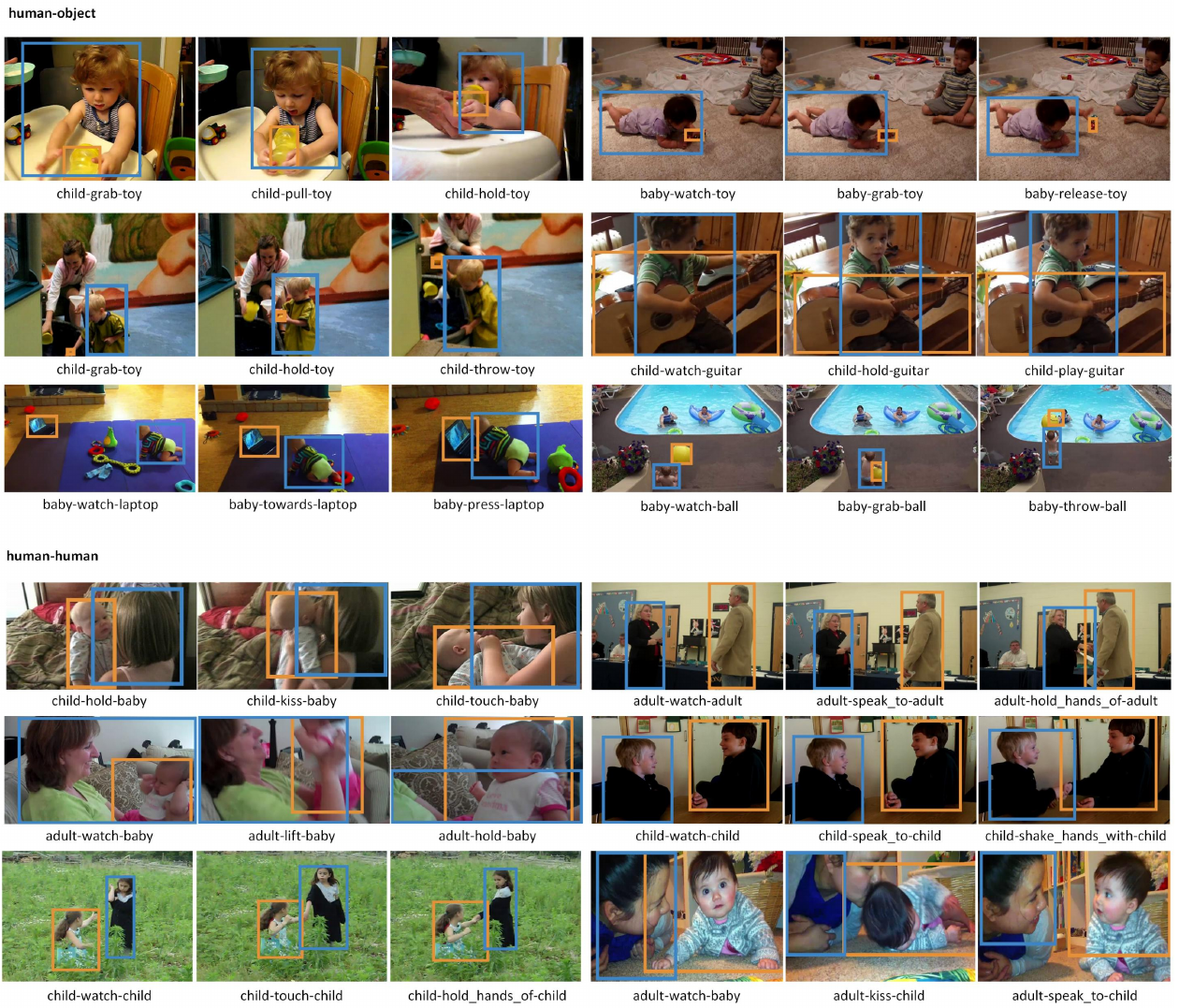}
	\end{center}
	\caption{Examples of the SemCo-VidOR datasets. Selected video frames of \textit{Human-Object} and \textit{Human-Human} interactions are showed.}
	\label{fig:example2}
\end{figure*}
\begin{figure}[t]
	\begin{center}
		\includegraphics[width=0.48\textwidth]{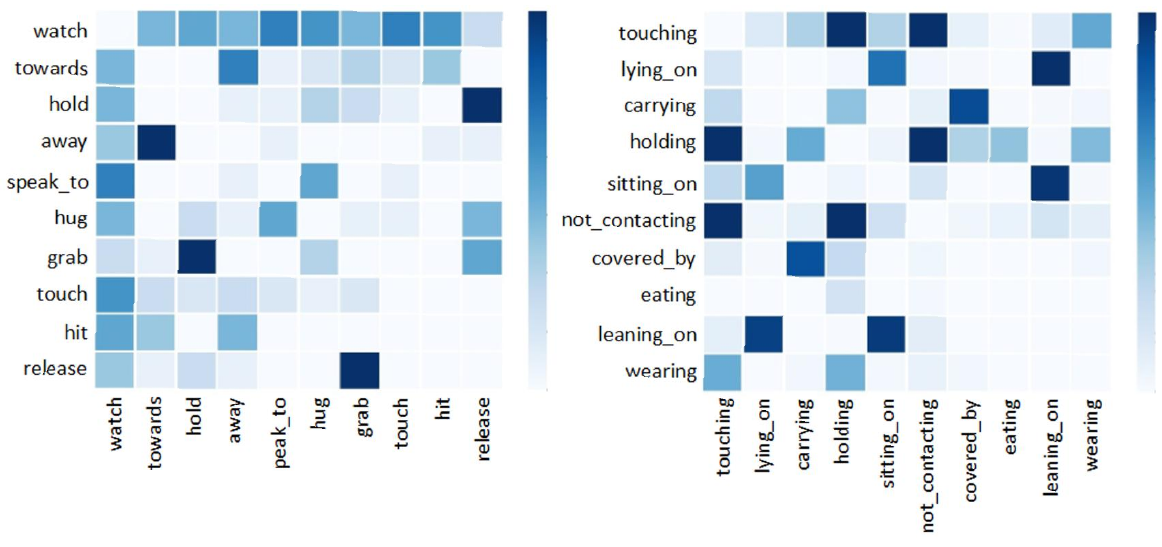}
	\end{center}
	\caption{Statistical distribution of relation transition in SemCo-VidOR (left) and SemCo-AG datasets (right).}
	\label{fig:datasettrans}
\end{figure}
\subsection{Forecasting and Reasoning Methods}
According to tasks, current future prediction methods are designed to achieve different forecasting goals. 
For the video forecasting task, generative models are required to predict future frames from intermediate representations. For instance, Luc \textit{et al.}~\cite{luc2017predicting} proposed an autoregressive convolutional neural network to learn scene dynamics by iteratively generating multiple frames. Alahi \textit{et al.}~\cite{alahi2016social} proposed an LSTM model to learn general human movement for predicting their future trajectories by taking into account the commonsense rules and social conventions. 

To forecast future states, a wide variety of methods have been developed. For example, Chen \textit{et al.}~\cite{sun2019relational} focused on multi-person action forecasting in videos and proposed a recurrent graph named Discriminative Relational Recurrent Network (DR$^2$N) for temporal interaction modeling. Wang \textit{et al.}~\cite{wang2021ttpp} repurposes a Transformer-style architecture to aggregate observed features, and then leverages a light-weight network to progressively predict future features and actions. Qiu \textit{et al.}~\cite{qiu2024multivariate} propose GoMMC for interpretable dense action anticipation, which can capture the influence between various objects and their interactions, allowing for a probabilistic selection of actions performed in the long term. 
Furthermore, some visual reasoning methods also focus on capturing the dependencies among objects~\cite{park2025swinlip}. For example, Jiang \textit{et al.}~\cite{jiang2021predicting} proposed a deep neural network model that integrates visual attention and hand position cues to predict the next active object. Oba~\cite{oba2025r2} propose an image-based retrieval method to retrieve the nearest neighbor motion in inference. For temporal relation reasoning, Rohit \textit{et al. }~\cite{girdhar2019video} proposed a video action transformer network to capture the temporal dependencies in videos for action recognition. Among them, some existing efforts focus on the combination of graph structure and a transformer network~\cite{rong2020self}.

To achieve scene graph anticipation, some relation-level reasoning methods have been developed. Peddi \textit{et al.}~\cite{peddi2024towards} infer observed video frames using object-centered associative representations and model the evolution of relationships between objects. Nguyen \textit{et al.}~\cite{nguyen2024hyperglm} propose the HyperGLM, a large-scale multimodal language model based on scene hypergraphs, designed to enhance reasoning about multi-directional interactions and higher-order relationships. Peddi \textit{et al.}~\cite{peddi2025towards} effectively mitigates bias issues in spatio-temporal scene graph generation and prediction by combining loss masking with coarse learning.

The VRF relies on capturing and modeling the semantic coherence in object interactions, as it underpins the evolution of events and scenes in videos. Existing methods neglect the interaction between visual cues and the semantic cues inherent in subject-object pairs, and their frame-by-frame modeling approach leads to overfitting of invariant relationships across consecutive frames. Different from those mentioned above, a method for VRF is supposed to capture object interactions for fine-grained reasoning. To achieve that, we propose a semantic coherent transformer method that augments the visual and semantic representation of the object interactions and utilizes a sparse encoding transformer to achieve frame-level reasoning.
\begin{figure*}[t]
	\begin{center}
		\includegraphics[width=\textwidth]{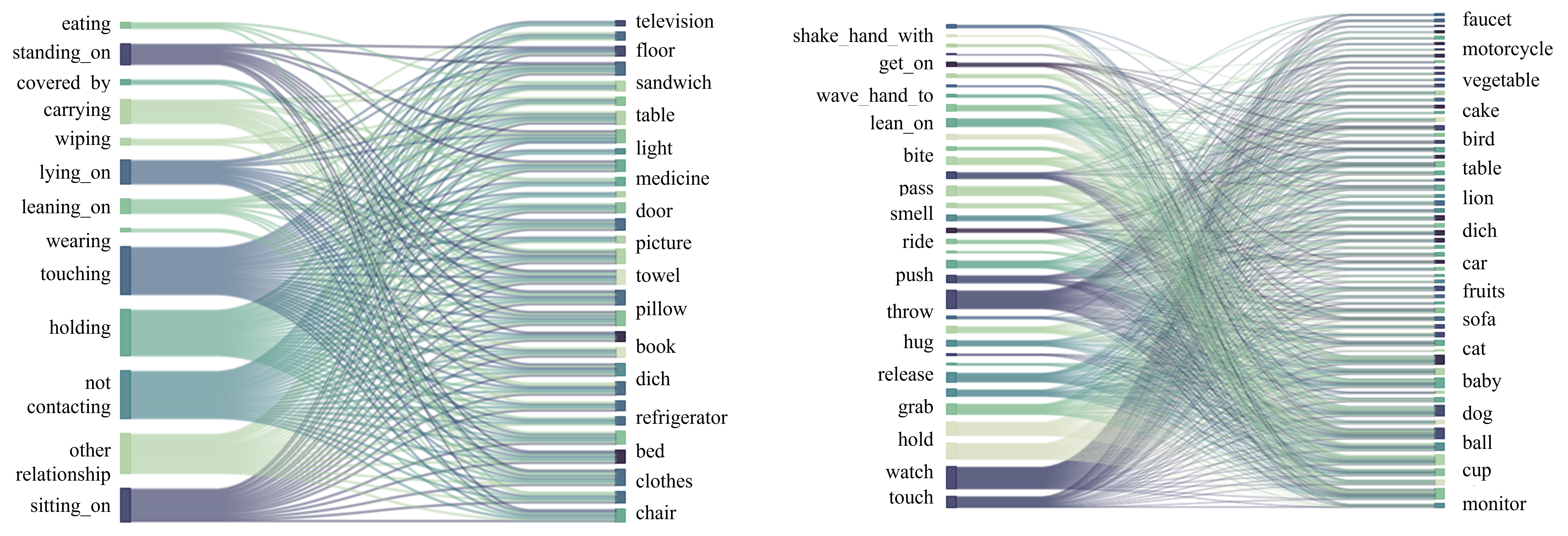}
	\end{center}
	\caption{Object-Relation Interaction of SemCo-AG dataset (left) and SemCo-VidOR dataset (right). The left side of each subfigure represents the relationship categories in the dataset, and the right side of the subfigure represents the object categories in the dataset. {Note that due to limited space, only some of the category names are shown in the figure.}}
	\label{fig:interaction}
\end{figure*}

\section{VRF DATASETS}
\label{sec:datasets}
The proposed VRF benchmark consists of two sub-datasets, \textit{i.e.} SemCo-AG and SemCo-VidOR. They are established based on Action Genome~\cite{ji2020action} and a video visual relationship detection dataset~\cite{shang2017video}, respectively. \par 
\subsection{Dataset Construction}
\label{sec:dataset_construction}
\textit{1) Predicate Vocabulary Generation.} \par
We follow three rules to generate the predicate vocabulary. First, we collect data with semantic relations and filter out spatial predicates (\textit{e.g.}, above, beneath). Second, some categories that cannot be accurately judged from visual information are filtered out (\textit{e.g.}, squeeze). Finally, to reduce the influence of long-tail distribution, predicate categories that account for less than 1\% of the total are deleted (\textit{e.g.}, twisting).\par 
\textit{2) Video and Keyframe Selection.} \par
The constant change of the predicate is regarded as a prerequisite for reasoning in VRF. Therefore, we select the data containing more than three predicate category annotations for a specific subject-object pair in the video. Moreover, if predicate annotations are repeated consecutively, we merged multiple keyframes with unchanged annotations into 2 keyframes. In VRF task, keyframes are not uniformly sampled from video frames. Therefore, in theory, the state of the two objects can be maintained for a long time in non-keyframes. Videos with more than 30 keyframes are truncated. \par 
\textit{3) Cleaning and Denoising.} \par
Excessively long timing video clips will cause great difficulty in visual relationship forecasting. By screening the length of video clips, videos with more than 30 keyframes are truncated. On the other hand, we try to filter videos and annotations to balance the number of predicate categories in the dataset which would alleviate the impact of the biased data distribution. \par 
\subsection{Dataset Statistics}
\label{sec:dataset_statistics}
\textit{1) Object Categories.} \par
Figure~\ref{fig:dataset} (a) and (b) show the log-distribution of object categories in SemCo-AG and SemCo-VidOR datasets. There are 30 object categories in SemCo-AG dataset and 64 object categories in SemCo-VidOR dataset. In particular, there are various subject categories besides \emph{person} in SemCo-VidOR dataset. Overall, compared with SemCo-AG dataset, the SemCo-VidOR dataset has more object categories, increasing the data diversity. \par 
\textit{2) Predicate Categories.}  \par 
Figure~\ref{fig:dataset} (c) and (d) show the distribution of relation categories in SemCo-AG and SemCo-VidOR datasets. There are 13 predicate categories in SemCo-AG dataset and 35 predicate categories in SemCo-VidOR dataset. Each video is annotated by a pair of subject-object and a time-series of predicates. Since the predicate number in SemCo-VidOR is more than that in SemCo-AG, SemCo-VidOR dataset is considered more challenging. \par 
\textit{3) Video Length.} \par 
Figure~\ref{fig:datasettime} shows the distribution of the number of keyframes per video in the proposed two datasets. As it shows, the overall distribution is close to a Gaussian distribution. The lengths of videos are concentrated in 4-12 keyframes, and the longest length is 30 keyframes. In this paper, we prepare 3 subsets for each dataset, containing 5, 10, and 15 keyframes respectively. \par 
\textit{4) Training and Test Sets.} \par 
To enrich the variability of the dataset, we adopt different train-test split methods for the two datasets. For SemCo-AG dataset, 50 videos for each relation category with a total of 650 videos are selected for the test set and the remaining 12797 videos are used to form the training set. For SemCo-VidOR dataset, we randomly select 400 videos for the test set with at least one sample from each category. The remaining 1523 videos are prepared for the training set. \par 

\subsection{Dataset Characteristics}
\label{sec:dataset_characteristics}
\textit{1) Example.} \par 
Figure~\ref{fig:example1} and Figure~\ref{fig:example2} shows annotation examples of the proposed datasets. We can find that for a specific $\textless$baby-toy$\textgreater$ pair or action annotation, the possible predicate sets are similar regardless of different visual appearance. For SemCo-AG dataset, Figure~\ref{fig:example1} shows the predicates between person and something in action \textit{picking up something} follow the series of $not\_contacting \Rightarrow touching \Rightarrow holding$. For SemCo-VidOR dataset, the $\textless$child-toy$\textgreater$ examples may have the predicate series of $watch \Rightarrow grab \Rightarrow hold \Rightarrow release$. \par 
\textit{2) Relation Transition.} \par 
Figure~\ref{fig:datasettrans} also shows the statistical distribution of relation transitions in SemCoBench. We calculate the number of all relationship transitions and normalize them within the same class of relation. In SemCo-VidOR dataset, the weights of $grab \Rightarrow hold$ and $hold \Rightarrow release$ are relatively high. The relationship category $watch$ may transfer to many kinds of relationship categories and those may also transfer to $watch$. In SemCo-AG dataset, $not\_contacting$ is more likely to transfer to $touching$ and $holding$, while $touching$ also tends to transfer to $holding$ and $not\_contacting$. The probability of transferring from $holding$ to $wearing$ is also in the top three.  Figure~\ref{fig:datasettrans} proves the possibility and significance to learn the relation transitions in videos. \par 
\textit{3) Object-Relation Interacion.} \par 
Figure~\ref{fig:interaction} shows the object-relation interactions of the proposed two datasets. The left side of each figure represents the relationship categories in the dataset, and each right side of the figure represents the object categories in the dataset. The object-relation interactions are complicated, which also indicates that the task of predicting the video visual relationship is difficult. Take SemCo-AG dataset as an example, a large number of relationship categories, such as \textit{holding} and \textit{touching}, interact with almost all objects. A small number of relationship categories, such as \textit{eating}, also interact with at least four objects (\textit{i.e., sandwich}, \textit{medicine}, \textit{groceries} and \textit{food}). The number of objects of each category is relatively balanced. \par 
\begin{figure*}[!t]
	\begin{center}
		\includegraphics[width=0.98\textwidth]{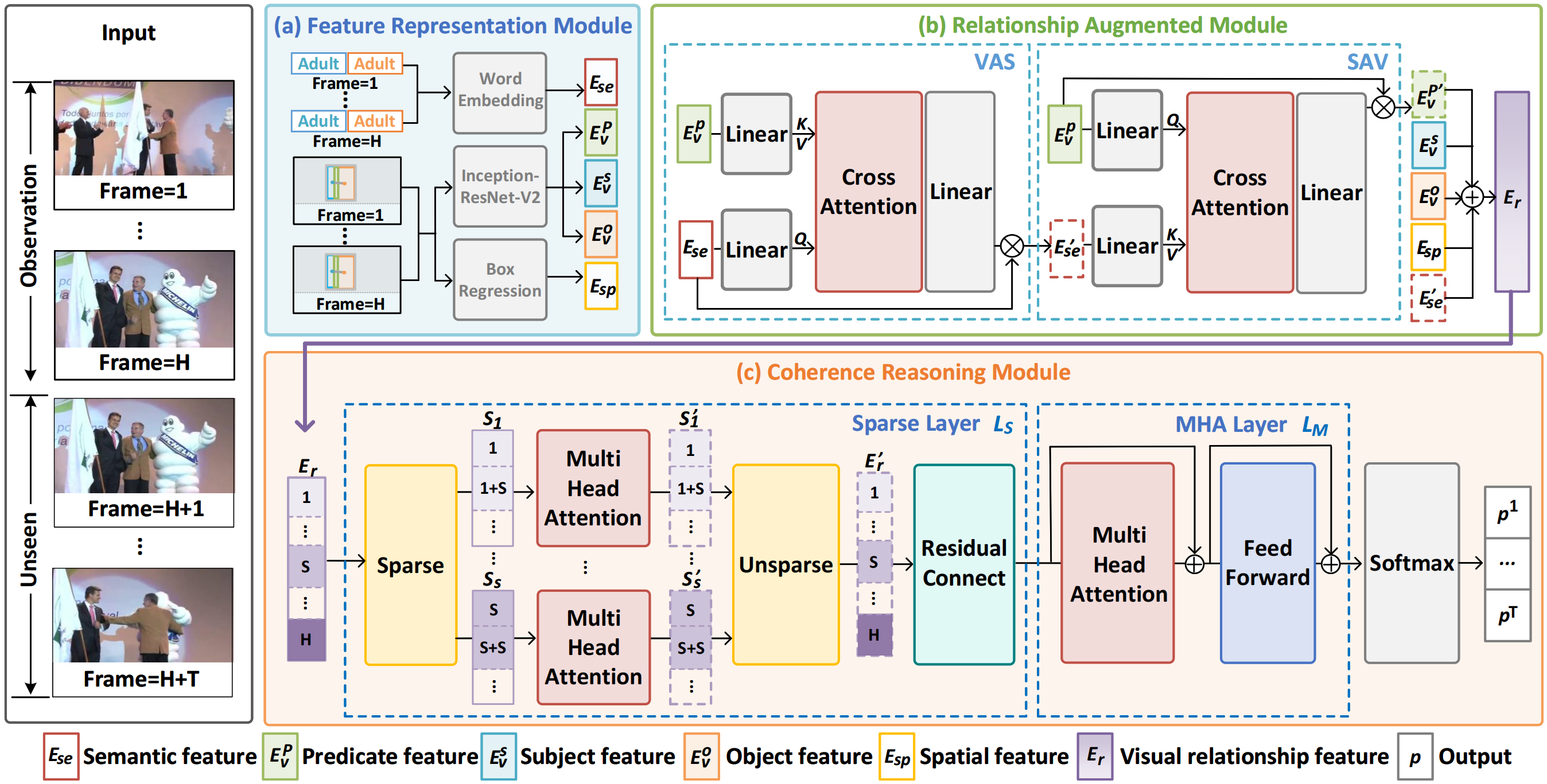}
	\end{center}
	\caption{The pipeline of the semantic coherent transformer method can be divided into three parts: feature representation module, relationship augmented module, and coherence reasoning module. In the feature representation module, visual feature, semantic feature, and spatial feature are embedded to represent the visual scene. Then, we designed a Relationship Augmented Module that bidirectionally augmented the visual representation and semantic representation of relationship predicates. Finally, for the Coherence Reasoning Module, we utilize sparse coding to model cross-frame relational transitions.}
	\label{fig:framework}
\end{figure*}
\section{TASK DEFINITION}
\label{sec:task_definition}
VRF aims to anticipate future interactions between a defined subject-object pair. More specifically, let $O \in \mathbb{R}^{N_{1}} $ and $P \in \mathbb{R}^{N_{2}}$ denote the object set and predicate set, respectively. And the $N_{1}$, $N_{2}$ are the number of object classes and relationship classes. The relationship set $R$ can be defined as $R = \{r(s,p,o)|s,o \in O,p \in P\}$, where $s$, $p$ and $o$ are respectively the subject, predicate, and object in a relationship triplet $r(s,p,o)$.  VRF aims to compute the probability: $ \mathbf{P}= \mathbf{P}\left(p^{0: T} \mid V^{-H: 0}, p^{-H: 0}, s,o \in O, p \in P \right)$, where $V^t$ is the frame at time $t$. $t=0$ represents the present. $V^{-H: 0}$ denotes the visual history of $H$ previous key frames. $p^{0: T}$ and $p^{-H: 0}$ represent the predicates in the future $T$ key frames and the past H frames, respectively. \par

\section{A VRF BASELINE METHOD}
\label{sec:method}
\subsection{Overview}
\label{sec:overview}

The semantic coherent transformer method can be divided into three parts: Feature Representation Modules (FRM), Relationship Augmented Modules (RAM), and Coherence Reasoning Module (CRM). 
The FRM is constructed to represent visual scenes, which embed visual features, semantic features, and spatial features.  
For FRM, video frames are fed as input, after feature extraction, the output consists of subject visual features $E^{s}_v \in \mathbb{R}^{ H\times D_{vis} }$, object visual features $E^{o}_v \in \mathbb{R}^{ H\times D_{vis} }$, predicate visual features $E^{p}_v \in \mathbb{R}^{ H\times D_{vis} }$, semantic features $E_{se} \in \mathbb{R}^{H\times D_{se} }$, and spatial features $E_{sp} \in \mathbb{R}^{ H\times D_{sp} }$. Among them, $D_{vis}$, $D_{se}$, and $D_{sp}$ represent the dimensions of visual features, semantic features, and spatial features, respectively. 
The RAM  is constructed to augment the visual representation and semantic representation of relational predicates. 
For RAM, $E^{p}_v$ and $E_{se}$ are the inputs of the local attention unit. These inputs are augmented in the first stage to update the semantic feature $E^{\prime}_{se} \in \mathbb{R}^{ H\times D_{se} }$, and then augmented in the second stage to update the visual feature $E^{p\prime}_v \in \mathbb{R}^{H\times D_{vis} }$. Finally, concatenate $E^{s}_v$, $E^{p\prime}_v$, $E^{o}_v$, $E^{\prime}_{se}$ and $E_{sp}$ to output the visual relationship feature $E_{r} \in \mathbb{R}^{H \times D_{model}}$. The $D_{model}$ denotes the dimension of the hidden feature in the transformer. 
The CRM is constructed to model cross-frame relational transitions. 
For CRM, $E_{r}$ are the inputs of the local attention unit. The visual relationship feature $E_{r}$ is refined and updated to $C \in \mathbb{R}^{H \times D_{model}}$. \par

\subsection{Feature Representation Module}
\label{sec:frm}
A single feature cannot represent the complex relationship between pairwise objects. In this paper, visual appearance, spatial feature, and semantic embedding are considered in the feature representation module, as shown in Figure~\ref{fig:framework} (a). \par 
\textit{1) Visual Feature $E_v$}\par
For a relationship instance $r(s,p,o)$, the $b_{s}$, $b_{so}$, and $b_{o}$ denote the bounding box of its corresponding subject, predicate, and object, respectively. Note that $b_{so}$ refers to the union of $b_{s}$ and $b_{o}$. We adopt Inception-ResNet-V2 as the backbone and extract the features of $b_{s}$, $b_{so}$ and $b_{o}$ from the fully connected layers, denoted as $E^{s}_{v}$, $E^{p}_{v}$, $E^{o}_{v}$, respectively. \par 
\textit{2) Spatial Feature $E_{sp}$} \par 
To get the relative spatial feature of bounding boxes, we adopt the idea of box regression~\cite{Hu2019Neural}.  Assume that $ \Delta\left(b_{s}, b_{o}\right)$ denotes the box delta that regresses the bounding box $b_{s}$ to $b_{o}$. Then $\operatorname{dis} \left(b_{s}, b_{o}\right) $ and $\operatorname{iou} \left(b_{s}, b_{o}\right) $ denote the normalized distance and IoU between $b_{s}$ and $b_{o}$. The union region of $b_{s}$ and $b_{o}$ is denoted as $b_{sp}$. The relative spatial location of subject and object can be defined as:

\begin{equation}
\begin{aligned}
E_{sp}=\left[\Delta\left(b_{s}, b_{o}\right) ; \Delta\left(b_{o}, b_{so}\right) ; \Delta\left(b_{o}, b_{so}\right)\right.\\\left.\operatorname{iou}\left(b_{s}, b_{o}\right) ; \operatorname{dis}\left(b_{s}, b_{o}\right)\right]. \end{aligned}
\end{equation} \par 
\textit{3) Semantic Feature $E_{se}$} \par 
We adopt a semantic embedding layer to map the object category into a word embedding. Note that the parameters of object categories are initialized with the pre-trained word representations, such as word2vec~\cite{mikolov2013efficient}.
\subsection{Relationship Augmented Modules} \label{sec:ram}  

Robust visual relationship features enable models to distinguish similar relationships, laying the basis for subsequent reasoning. Visual relationship features inherently contain rich multi-modal cues that provide valuable information. The visual cues provide fine-grained scene details, including object appearances, spatial layouts, and local visual patterns. The semantic cues provide high-level contextual knowledge, such as the meaning of object categories. Existing methods leverage visual features from subject-object pairs to infer future relational transitions, resulting in an inability to distinguish between visually similar predicate relationships. In fact, the interaction between visual cues and the object's inherent semantic cues can overcome this limitation.

We designed a module that leverages the mutual augmentation of visual and semantic features to yield more robust visual relationship features, as shown in Figure~\ref{fig:framework} (b). The module involves two steps. The first step is the Visual-Augment-Semantic (VAS), where we use the visual features $E^{p}_{v}$ to augment the semantic features $E_{se}$ of the corresponding frames and capture their temporal information simultaneously. The second step is the Semantic-Augment-Visual (SAV), where we use the semantic features $E_{se}^{\prime}$ obtained in the first step to augment the visual features of the corresponding frames. 

\subsubsection{Visual-Augment-Semantic}
We augment semantic features with visual features at each time step to ensure that the semantic representations are dynamically updated based on video content. To effectively capture such temporal dependencies, we adopt a Transformer~\cite{Vaswani2017Attention}, which has been widely used in temporal modeling tasks. In this framework, self-attention compares a feature to all others in the sequence by projecting them into Query ($Q$), Key ($K$), and Value ($V$) embeddings. The output for the model is computed as an attention-weighted sum of values $V$, with the attention weights obtained from the product of the query $Q$ with keys $K$. A location embedding is also added to incorporate positional information which is lost in the non-convolution setup. 

This augmentation process is performed using a multi-head attention mechanism, where the semantic features $E_{se}$ serve as the query and the predicate visual feature $E^{p}_{v}$ serve as the key and value. 
\begin{equation}
\begin{split}
    Q = E_{se}W^{Q},\quad  K= E^{p}_{v}W^{K},\quad  V= E^{p}_{v}W^{V},
\end{split}
\end{equation}
where $W^{Q}$, $W^{K}$, $W^{V}$ are learnable parameter matrices. Then, multi-head attention layers are preformed to get the augmented  semantic feature $E_{se}^{\prime}$:
\begin{equation}
\label{eq:vas}
\begin{split}
&E_{se}^{\prime} = \operatorname{MultiHead}\left(Q,K,V\right),
\end{split}
\end{equation}

The semantic features augmented by VAS not only distinguish objects but also differentiate between different interaction patterns, as visual feature provide contextual information for semantic feature within video content.

\subsubsection{Semantic-Augment-Visual}
We augment visual features with semantic features, where the VAS refines these semantic features at each time step, enabling the visual representation to be dynamically updated according to the object categories. This augmentation process is performed using a multi-head attention mechanism, where the visual feature $E^{p}_{v}$ serves as the query and the semantic feature $E_{se}^{\prime}$ serves as the key and value:
\begin{equation}
\begin{split}
    Q = E^{p}_{v}W^{Q},\quad  K= E_{se}^{\prime}W^{K},\quad  V= E_{se}^{\prime}W^{V},
\end{split}
\end{equation}
\begin{equation}
\label{eq:sav}
\begin{split}
& E^{p \prime}_{v} =\operatorname{MultiHead}\left(Q,K,V\right),
\end{split}
\end{equation} 
where $E^{p \prime}_{v}$ represents the visual features augmented by SAV. 

By injecting the augmented semantic cues back into the visual feature, visual features focus more on regions relevant to the object's relationship, as category labels provide explicit semantic constraints.

After obtaining the augmented predicate visual feature $E^{p \prime}_{v}$ and the augmented semantic feature $E_{se}^{\prime}$, we concatenate them with the subject visual feature $E^{s}_{v}$, object visual feature $E^{o}_{v}$, and the spatial feature $E_{sp}$ to form the final robust visual relationship feature $E_{r}$, which serves as the foundation for modeling the semantic coherence of relationship changes:
\begin{equation}
\label{eq:concat}
    E_{r} = Concat\big(E^{s}_{v},\, E^{p \prime}_{v},\, E^{o}_{v},\, E_{se}^{\prime},\, E_{sp}\big).
\end{equation} \par 
By concatenating all feature representations, different aspects of visual relationships—subject, object, predicate, semantic context, and spatial layout—can be integrated into a unified representation, enabling subsequent modules to perform joint inference on multiple information sources.

\subsection{Coherence Reasoning Module}
\label{sec:fsr}
\subsubsection{Sparse Coding}
Reasonable relational shifts that correspond to semantic coherence typically require traversing a sequence of visual relationships. This sequence contains both relationships that undergo transitions and relationships that remain unchanged. For example, in the action of `\textit{Opening the refrigerator}', the relationship of $\textless$\textit{person-holding-refrigerator}$\textgreater$ is continuous throughout the video frames. Existing methods employ a frame-by-frame modeling approach to learn object interactions across consecutive frames. However, such an approach causes the model to focus excessively on consecutive frames where relationships remain unchanged, thereby neglecting frames where relationships undergo transition.

To address this issue, we applied $L_{S}$ layers sparse coding, where each sparse coding layer undergoes both sparse and unsparse processes. As shown in Figure~\ref{fig:framework}(c), this enables the model to learn changes in visual relationships between non-consecutive video frames, making it adept at capturing relational shifts and thereby better understanding the semantic coherence of object interactions.

\textit{1) Sparse Process} \par
In particular, the visual relationship feature $E_{r}$ is the input of this module, which is divided into $H$ blocks. $H$ denotes the number of observed frames. Following $\operatorname{Sparse}(\cdot)$ operations~\cite{stevens2025sequential, sun2024mc}, the $H$ blocks are grouped into $s$ sets, blocks within each set originally separated by an interval of $s$. $\operatorname{Sparse}(\cdot)$ operations can be specifically represented by combining the $i$-th block, the $(i+s)$-th block, the $(i+2 \times s)$-th block... up to the $(i+(H\div s-1)\times s)$-th block of $E_{r}$ into $S_{i}$. The specific operation of $\operatorname{Sparse}(\cdot)$ can be expressed as:
\begin{equation}
\begin{split}
\operatorname{Sparse}(E_{r})&=\operatorname{Concat}(E_{r}^{i}, E_{r}^{i+s}, E_{r}^{i+2\times s},..., E_{r}^{i+(H\div s-1)\times s})\\&=S_{i}, \quad {i} \in [1,s],
\end{split}
\end{equation}

Flattening these $s$ sets of features yields feature subsequences $S_i$, where $i$ ranges from 1 to $s$. This process can be formulated as:
\begin{equation}
\begin{split}
S_{i}=\operatorname{Flatten}(\operatorname{Sparse}(E_{r}) ),\quad {i} \in [1,s],
\end{split}
\end{equation}

After sparse $E_{r}$ into $s$ feature subsequences $S_{i}$, we apply self-attention to each subsequence, where $S_{i}$ simultaneously serve as the query $Q_{i}$, key $K_{i}$, and value $V_{i}$, where $i \in [1,s]$:
\begin{equation}
    \begin{split}
        Q_i = S_i W^{Q}_{i},\quad  K_{i}= S_i W^{K}_{i},\quad  V_{i}= S_i W^{V}_{i},\quad{i} \in [1,s],
    \end{split}
\end{equation}
where $W^{Q}_{i}$, $W^{K}_{i}$, $W^{V}_{i}$ are learnable parameter corresponding to each subsequence $S_i$.

Enable the model to learn which relationship transitions better correspond to global semantic coherence and should be assigned higher weights. This process can be formulated as: 
\begin{equation}
\begin{split}
{S_{i}}^{\prime} =\operatorname{MultiHead}\left(Q,K,V\right), \quad{i} \in [1,s],
\end{split}
\end{equation}

Each sparse block $S_{i}$ obtains its corresponding self-attention result $S^{\prime}_{i}$. 

\textit{2) Unsparse Process}\par
We adopt the unsparse strategy $\operatorname{Unsparse(\cdot)}$, which is the reverse of the sparse strategy, to aggregate the self-attention results of $s$ feature subsequences into one and flatten it: 
\begin{equation}
\begin{aligned}
U=\operatorname{Flatten}(\operatorname{Unsparse}(S^{\prime}_{1},S^{\prime}_{2},...,S^{\prime}_{s})),
\end{aligned}
\end{equation}

Establish residual connections for the flattened output $U$. This approach preserves original feature information while enhancing training stability in deep structures, mitigating gradient vanishing issues, and promoting effective feature fusion. The process can be expressed as: 
\begin{equation}
E^{\prime}_{r}=E_{r}+\operatorname{DropPath}(\gamma_{1} \cdot U),
\end{equation}
where ${\gamma}_{1}$ is a learnable scalar parameter, $E^{\prime}_{r}$ is the output feature processed by the sparse and unsparse. 
During training, by randomly discarding residual paths with probability $p$, and scaling the retained portion by $\frac{1}{1 - p}$, $\operatorname{DropPath(\cdot)}$ enhances the model's regularisation capability without altering the expected output. The specific operation of $\operatorname{DropPath(\cdot)}$ is as follows: 
\begin{equation}
\begin{aligned}
\operatorname{DropPath}(\gamma_{1} \cdot U)=m \cdot \frac{1}{1 - p} \cdot(\gamma_{1} \cdot U), 
\end{aligned}
\end{equation}
where $p$ denotes the discard probability, and $m$ represents a random variable following a Bernoulli distribution, indicating whether the current path is retained.

\subsubsection{Multi-head Transformer}
Finally, we apply $L_{M}$ layers of multi-head attention to obtain the final embedding $C$, where the query, key and value are the output features $E^{\prime}_{r}$ produced by the sparse coding. In this way, the model can reintegrate the visual relational features across the entire video sequence, ensuring the model preserves semantic coherence in relationship transitions during the entire action event. This process can be formulated as:
\begin{equation}
    \begin{split}
        Q&=E^{\prime}_{r}W^{Q},\quad K=E^{\prime}_{r}W^{K},\quad V=E^{\prime}_{r}W^{V}\\
    \end{split}
\end{equation}
\begin{equation}
\begin{split}
C&=\operatorname{MultiHead}(Q,K,V),
\end{split}
\end{equation}
\par
The final decoder embedding is sent to an FC layer and a softmax layer to get the final prediction $p$:
\begin{equation}
   p = \operatorname{Softmax}(\operatorname{Linear}(C)),
\end{equation}\par
After the computation of SemCoFormer, the output of the method $p$ is extracted for an end-to-end optimization by calculating the cross-entropy loss: 
\begin{equation}
\mathcal{L} = -\frac{1}{T} 
\sum_{t=1}^{T} 
\log \, p^{t}_{\hat{y}_{t}} ,
\end{equation}
where $T$ is the number of future key frames. The predicted probability distribution is denoted as $p^{t}$, and $\hat{y}_{t}$ is the target probability distribution.

\section{EXPERIMENTS}
\label{sec:experiments}
\subsection{Evaluation Methods}
\label{sec:evaluation_methods}
We train and evaluate our models on the aforementioned train-test split of SemCo-VidOR and SemCo-AG datasets. The evaluation metric is the Accuracy (Acc) and mean Average Precision (mAP). mAP is the mean of the precision at different recall rates, which measures the model's overall performance in multi-category tasks by calculating the average of the area under the precision-recall curves of the various categories. Acc focuses on the proportion of correctly classified samples to the total number of samples as a reflection of the overall classification accuracy. \par 
\subsection{Experimental Methods}
\label{sec:experimental_methods}
We evaluate four types of baselines on the proposed VRF benchmark: \par 
\textit{1) Baselines methods in probabilistic model}\par 
\begin{itemize}
    \item \textbf{PM}, which predicts the future predicates only based on the statistics on the probability of predicate transition in the dataset.
\end{itemize} \par 
\textit{2) Baseline methods in sequence modeling} \par 
\begin{itemize}
    \item \textbf{LSTM} \cite{hochreiter1997long}, which has been used as the advanced version of RNN network for long-term temporal modeling.
    \item \textbf{GRU} \cite{cho2014learning}, which is the simple variant of LSTM.
    \item \textbf{ST-GCN}~\cite{yan2018spatial}, which was proposed for natural language processing as a variant of GCN~\cite{kipf2016semi} and revealed the effectiveness of learning on graph-structured data.
    \item \textbf{Transformer} \cite{Vaswani2017Attention}, which computes the feature dependencies through self-attention mechanism.
\end{itemize} \par 
\textit{3) Baseline methods in visual relationship detection}\par 
\begin{itemize}
    \item \textbf{VR-LP}~\cite{Lu2016Visual}, which proposed a baseline method for visual relationship detection.
    \item \textbf{VtransE}~\cite{Zhang2017Visual}, which introduced translation embedding methods to visual relationship detection.
\end{itemize} \par 
\textit{4) Baseline methods in action/scene graph forecasting}\par 
\begin{itemize}
    \item \textbf{DR$^2$N}~\cite{sun2019relational}, which is a baseline for relational action forecasting based on graphs and GRU~\cite{cho2014learning}.
    \item \textbf{GTN}~\cite{rong2020self}, which integrated Message Passing Networks into Transformer-style architecture to build an expressive encoder for the informative representation of molecules.
    \item \textbf{SceneSayerSDE}~\cite{peddi2024towards}, which inferred representations of future relationships by solving a Stochastic Differential Equation.
    \item 
    \textbf{SemCoFormer} (Ours), which augments cross-modal representations and employs a sparse coding for coherence reasoning. 
\end{itemize}
\subsection{Standard Forecasting Setting}
\label{sec:strandard_forecasting_setting}
The results in Table~\ref{tab:results} show the performance of current baselines on the proposed SemCoBench. Note that we implemented some of the baselines according to their paper.  \par 
PM is only based on prior knowledge in datasets, which is unsatisfactory for responding to the challenges of real-world applications. Transformer outperforms the other sequence modeling methods in VRF, due to its powerful ability in capturing independencies among items. 
Compared with Transformer~\cite{Vaswani2017Attention}, we trained the model to distinguish similarity relationships using more robust visual relationship features.\par
\begin{figure*}[t]
	\begin{center}
		\includegraphics[width=\textwidth]{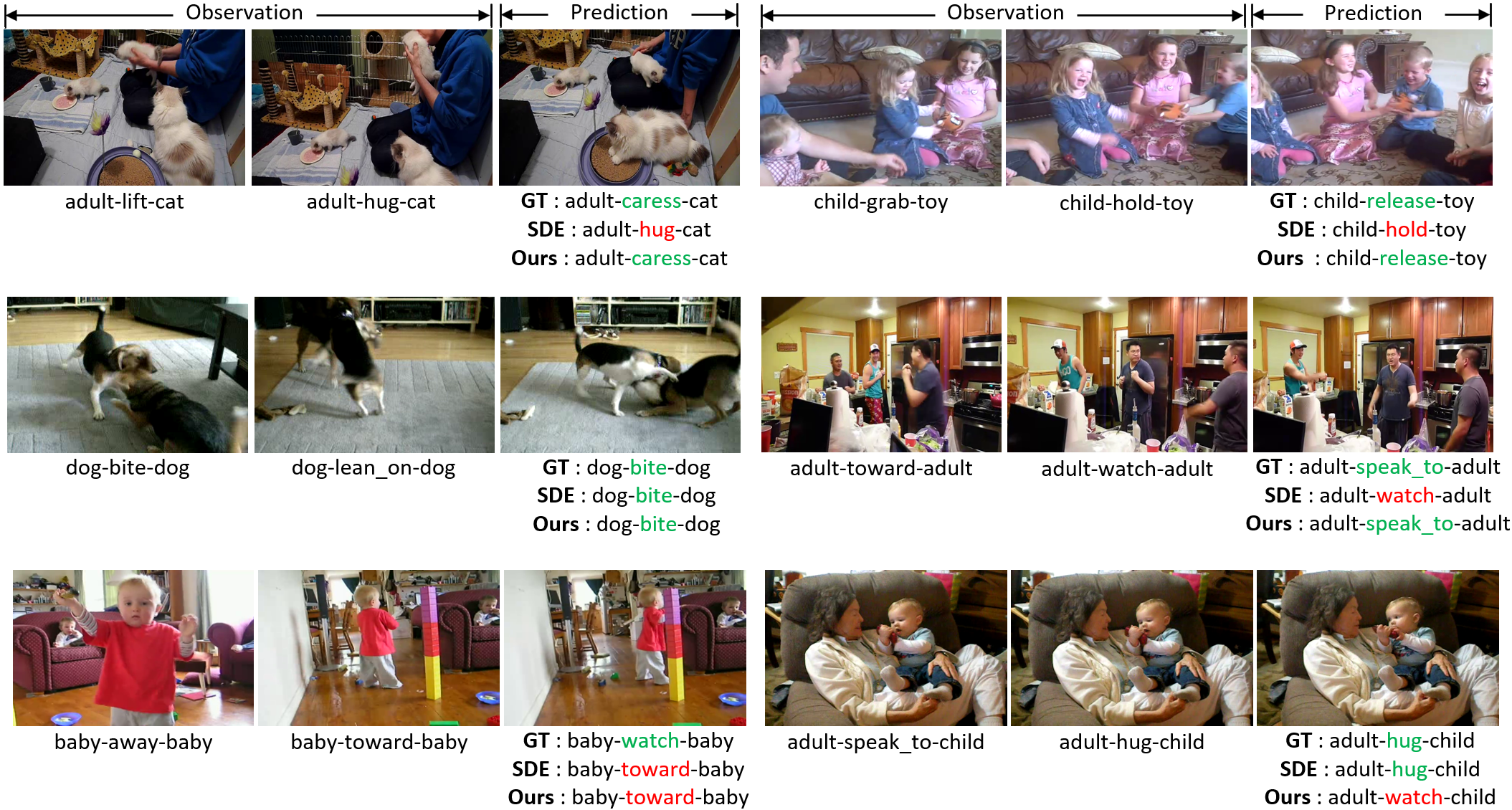}
	\end{center}
        \vspace{-5mm}
	\caption{The visualization results of SceneSayerSDE (SDE)~\cite{peddi2024towards} and SemCoFormer. The green color denotes the correct predictions while the Red color represents the false predictions.}
        \vspace{-5mm}
	\label{fig:result}
\end{figure*}
\begin{table}[t]
	\centering
	\caption{Comparison with the baseline methods in SemCo-VidOR dataset and SemCo-AG dataset. Note that for fair comparison, we added transformer to visual relationship detection methods, which are reported as VR-LP* and VtransE*.}
    \vspace{1mm}
    \scriptsize
    \resizebox{0.4\textwidth}{!}{
	\begin{tabular}{ccccc}
		\toprule
		\multirow{2}[4]{*}{} & \multicolumn{2}{c}{SemCo-VidOR} & \multicolumn{2}{c}{SemCo-AG} \\
		\cmidrule{2-5}          & Acc   & mAP   & Acc   & mAP \\
		\midrule
		PM & 17.25 &	8.51 &	15.85 &	10.08\\
		LSTM~\cite{hochreiter1997long} & 20.50 & 6.53 & 63.38 &70.68 \\
		GRU\cite{cho2014learning} & 28.25 & 19.37 & 67.08 & 73.90 \\
		ST-GCN\cite{yan2018spatial}  & 29.50 &	25.30 & 64.62 & 73.48  \\
		Transformer\cite{Vaswani2017Attention}& 33.50 &	25.87 & 71.69 &	77.87 \\
		{VR-LP\cite{Lu2016Visual}} & {25.25} & {20.74} & {62.46} & {67.78} \\
		{VtransE~\cite{Zhang2017Visual}} & {25.25} & {20.73} & {61.23} & {64.85}\\ 
		{VR-LP*} & {33.25} & {31.39} & {72.77} & {79.03} \\
		{VtransE*} & {34.25} & {26.97} & {71.38} & {78.84} \\
		{DR$^2$N~\cite{sun2019relational}} & {28.25} & {24.59} & {70.00} & {75.34} \\ 
		{GTN~\cite{rong2020self}}  & {32.25} & {28.36} & {72.00} & {77.98} \\
        SceneSayerSDE~\cite{peddi2024towards} & 34.47 & 30.07  & 73.89 & 78.56 \\
        \textbf{SemCoFormer(ours)} & \textbf{36.84} & \textbf{33.59}  & \textbf{75.50} & \textbf{80.94} \\
		\bottomrule
	\end{tabular}%
    }
	\label{tab:results}%
\end{table}%
Due to the lack of dynamic modeling ability, the visual relationship detection methods such as VR-LP~\cite{Lu2016Visual} and VtransE~\cite{Zhang2017Visual} are not competitive on the proposed SemCoBench. In order to achieve a fair comparison, we combine transformer and these two VRD baselines to improve dynamic modeling. 
Compared with the original VRD baselines (VR-LP~\cite{Lu2016Visual} and VtransE~\cite{Zhang2017Visual}) and the VRD baselines combined with Transformer (VR-LP* and VtransE*), the SemCoFormer achieves improvement on Acc and mAP for both methods. 
This further demonstrates that the detection-oriented approaches are not specifically designed to capture the temporal dynamics and contextual dependencies required for prediction. In contrast,our SemCoFormer model explicitly incorporates temporal dynamics modeling and contextual relational reasoning, which enables it to better capture the dependencies required for prediction, leading to consistently better performance. \par 
 
Both DR$^2$N~\cite{sun2019relational} and GTN~\cite{rong2020self} use graphs to model object interactions. GTN~\cite{rong2020self} is a widely used method that plugs graph structure into the Transformer. SceneSayerSDE~\cite{peddi2024towards} accomplishes this by representing nonlinear patterns, while SemCoFormer employs cross-modal augmented visual relationship features to model cross-frame dynamic relationships with sparse coding. For SemCo-VidOR dataset, SemCoFormer improved Acc by 2.37\% and mAP by 3.52\% over SceneSayerSDE~\cite{peddi2024towards}. For SemCo-AG dataset, SemCoFormer outperforms SceneSayerSDE~\cite{peddi2024towards} on Acc and mAP by 1.61\% and 2.38\%, respectively. 
SemCoFormer generally outperforms other methods on the proposed SemCoBench, primarily due to its ability to integrate cross-modal representations more effectively, resulting in stronger relational understanding. 

To analysis the performance of the proposed method SemCoFormer, Figure \ref{fig:result} shows the result of two methods: SceneSayerSDE (SDE)~\cite{peddi2024towards} and SemCoFormer for Visual Relationship Forecasting on SemCo-VidOR dataset and SemCo-AG dataset. 
SceneSayerSDE~\cite{peddi2024towards} inferred representations of future relationships by solving a Stochastic Differential Equation. In complete events, there will exist duration relationships that connect different actions. For example, on the right side of the first row in Figure~\ref{fig:result}, the \textit{hold} relationship connects the action of `\textit{Grab the toy}' and the action of `\textit{Release the toy}'. SceneSayerSDE~\cite{peddi2024towards}, however, overemphasizes duration relationships, leading to incorrect predictions of the $\textless$\textit{child-hold-toy}$\textgreater$ relationship in subsequent frames. SemCoFormer cleverly avoids overfitting duration relationships by modeling cross-frames, focusing instead on understanding the semantic coherence of relational dynamics. Additionally, when similar relationship transitions occur, SceneSayerSDE~\cite{peddi2024towards} become confused. For example, on the left side of the second line in Figure~\ref{fig:result}, the \textit{watch} relationship and the \textit{speak to} relationship are visually very similar because the subject and object do not change their relative positions, and both relationships could potentially occur simultaneously, leading the visual-based method SceneSayerSDE~\cite{peddi2024towards} to incorrectly predict the $\textless$\textit{adult-watch-adult}$\textgreater$ relationship. In contrast, the SemCoFormer method employs a relationship augmented module to obtain robust visual relational representations, enabling it to distinguish similar relations effectively and predict the correct relationship $\textless$\textit{adult-speak\_to-adult}$\textgreater$.

SemCoFormer can predict visual relationships that are more consistent with actual action trends. For example, in the first row on the left side of the Figure~\ref{fig:result}, after the adult \textit{lift} and \textit{hug} the cat, SemCoFormer can accurately predict the next relationship $\textless$ \textit{adult-caress-cat} $\textgreater$ in the adult's action of `\textit{Petting the cat}'. 
However, the anticipation of future object interactions are prone to be wrong when the probability of future occurrences is not unique or the annotations and visual information do not match uniquely. For example, in the left of the third row of Figure~\ref{fig:result}, after leaving another child, a child may toward him again, or may watch him. In the right third row of Figure~\ref{fig:result}, the ground truth annotations of the predicted frame match the visual information of future frames, but the visual relationships predicted by SemCoFormer also conform to the visual information.\par 

\begin{figure}[t]
	\begin{center}
		\includegraphics[width=0.48\textwidth]{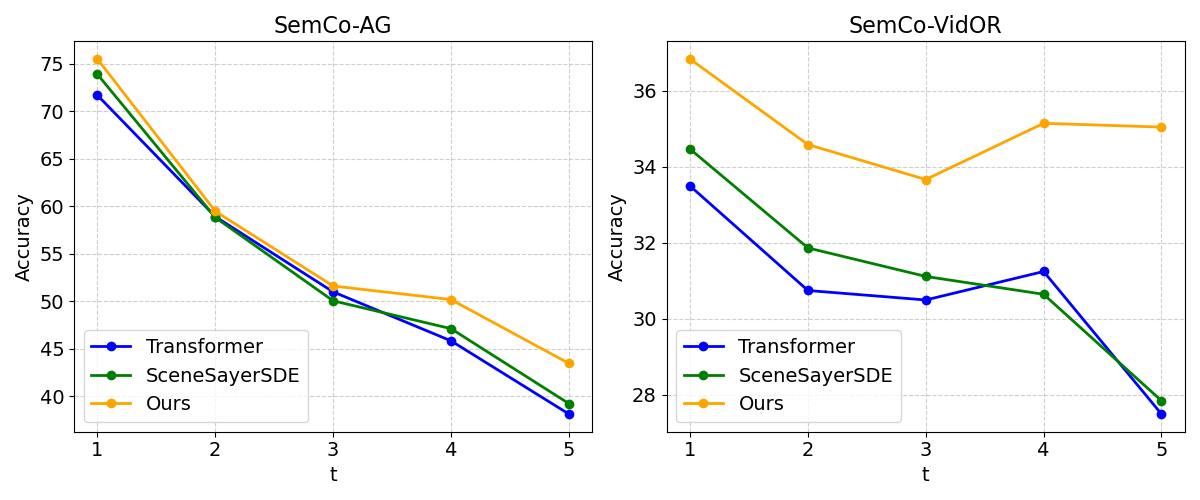}
	\end{center}
	\caption{sequence relationship forecasting results on SemCo-AG dataset and SemCo-VidOR dataset.}
	\label{fig:sequence_1}
\end{figure}
\begin{figure}[t]
	\begin{center}
		\includegraphics[width=0.48\textwidth]{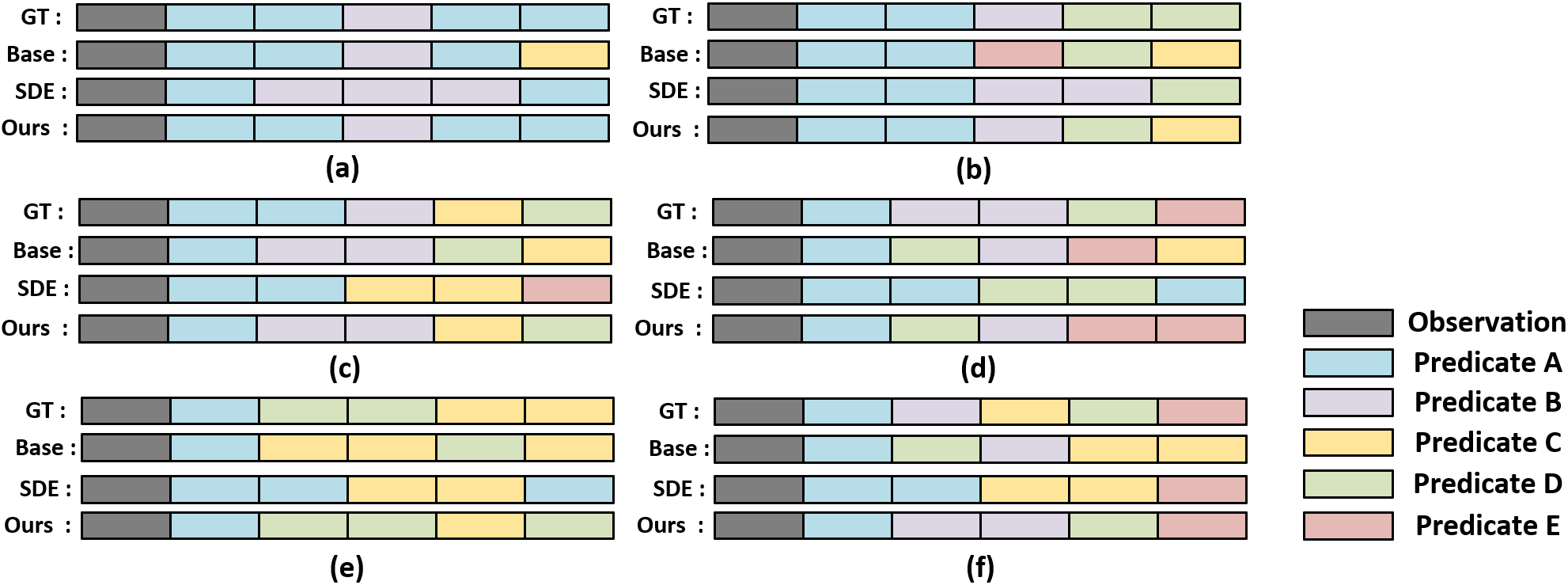}
	\end{center}
	\caption{The sequence forecasting results of Transformer (Base)~\cite{Vaswani2017Attention}, SceneSayerSDE (SDE)~\cite{peddi2024towards} and SemCoFormer (Ours). Every subgraph in the figure represents a sample. Every color denotes a type of predicate. Note that colors are not assigned fixed categories for different samples.}
	\label{fig:sequence}
\end{figure}
\subsection{Forecasting Setting Variants}
\label{sec:forecasting_setting_variants}
\subsubsection{Forecast Sequence Relationships}
We conduct a sequence forecasting experiment on SemCo-AG dataset and SemCo-VidOR dataset. The results are shown in Figure~\ref{fig:sequence_1}. $t$ represents the number of unknown keyframes in the future. Our first observation is that, as $t$ grows larger, the Acc and mAP decline accordingly. This is expected since when the testing sequence gets longer, the task would become more challenging and be with more uncertainty. In other words, the performance of relationship forecasting relies on the accurate observation of the past. 
However, our method clearly outperforms the baseline method (\textit{i.e.}, Transformer~\cite{Vaswani2017Attention}, SceneSayerSDE~\cite{peddi2024towards}) in its ability to predict the future over the long term. The possible reason is that our approach models cross-frame relational dynamics, extending beyond connections between neighboring frames. The SemCoFormer focuses on understanding the entire action event. Thus, even as the number of predicted future frames increases, SemCoFormer can still capture the semantic coherence of target interactions, achieving superior prediction results.

Figure~\ref{fig:sequence} shows some examples of sequence forecasting. The grey color represents the input sequence, and other colors represent the predicted labels. If the prediction and the label are in the same color, the prediction is correct, otherwise, it is wrong. The Transformer (Base)~\cite{Vaswani2017Attention}, SceneSayerSDE (SDE)~\cite{peddi2024towards} method is set as the comparison methods. 
Experimental results show that when the sequence grows, the prediction of the proposed SemCoFormer is more accurate, stable, and diverse than the comparison methods. Its predictions no longer merely replicate observed frames, but instead comprehend the semantic transformations of relational sequences within the video, thereby yielding more coherent prediction results. This advantage benefits from SemCoFormer’s cross-modal augmentation, which enhances the robustness of visual relationship representations. By performing consequent reasoning on this foundation, the model can better distinguish similar relationships, thereby predicting more accurate and diverse outcomes.

\subsubsection{Forecasting in Different Video Lengths}
To evaluate the model performance on different lengths of videos, we design a subtask that requires forecasting on videos with 5, 10, and 15 keyframes, respectively. For the 5-frame evaluation, videos with more than 5 keyframes are cut to generate the 5-frame subset. The 10-frame and 15-frame subsets are generated using the same methods as the 5-frame. Experimental results in Table \ref{tab:forecasting} show that the model performs better on the 10-frame subset in both datasets. That is mainly because the lengths of videos in our two datasets are centered in 10 keyframes. The other finding is that the model performs better on the 15-frame subset than the 5-frame subset in general, since the former provides more information.
\begin{table}[t]
	\centering
	\caption{Forecasting in Different Video Lengths on SemCo-AG dataset and SemCo-VidOR dataset.}
    \vspace{1mm}
    \resizebox{0.48\textwidth}{!}{
	\begin{tabular}{llcccccc}
        \toprule
        \multirow{2}{*}{Dataset} & \multirow{2}{*}{Method} 
        & \multicolumn{2}{c}{Length=5} & \multicolumn{2}{c}{Length=10} & \multicolumn{2}{c}{Length=15} \\
        \cmidrule(lr){3-4}\cmidrule(lr){5-6}\cmidrule(lr){7-8}
         &  & Acc & mAP & Acc & mAP & Acc & mAP \\
        \midrule
        \multirow{4}{*}{SemCo-AG} 
          & Transformer~\cite{Vaswani2017Attention} & 69.54 & 74.62 & 71.69 & 77.87 & 71.54 & 76.69 \\
          & SceneSayerSDE~\cite{peddi2024towards} & 70.98 & 76.78 & 73.89 & 78.56  &  71.63  &   77.24  \\
          & SemCoFormer  & 73.03 & 75.83 & 75.50 & 80.94 & 72.57 & 78.64 \\
        \midrule
        \multirow{4}{*}{SemCo-VidOR} 
          & Transformer ~\cite{Vaswani2017Attention} & 27.25 & 20.36 & 33.50 & 25.87 & 31.25 & 23.17 \\
          & SceneSayerSDE~\cite{peddi2024towards}&   30.83   & 25.12 & 34.47 & 30.07  &  33.27   &  27.78   \\
          & SemCoFormer  & 33.83 & 27.90 & 36.84 & 33.59 & 33.58 & 29.20 \\
        \bottomrule
    \end{tabular}}
	\label{tab:forecasting}%
\end{table}%

\subsection{Ablation Studies}
\subsubsection{Module Analysis}
We design ablation studies on SemCo-VidOR and SemCo-AG datasets to understand the importance of each component of our model. In module analysis, the relationship augmented module consists of two steps: the Visual Augment Semantic (VAS) and the Semantic Augment Visual (SAV). The importance of the VAS and SAV steps is analyzed through a step-by-step (VAS, SAV) and complete relationship augmented module (VAS+SAV). The Sparse Coding (SC) is added to the relationship augmented module to study the importance of the sparse coding. 

Results are presented in Table \ref{tab:module}, which indicate that the three components working together achieved the best performance on both datasets. Furthermore, the performance of VAS and SAV when used individually was inferior to their combined use. In addition, the VAS step is more important than the SAV step for SemCo-AG dataset. The VAS and SAV steps work to different extents on SemCo-AG and SemCo-VidOR datasets, respectively. This is due to the differing subject categories and scenarios present in SemCo-AG and SemCo-VidOR datasets. Sparse coding strategies demonstrated significant performance on both SemCo-AG and SemCo-VidOR datasets, confirming that sparse coding of cross-frames effectively preserves semantic coherence in object interactions, during visual relationship forecasting tasks.

\begin{table}[t]
	\centering
	\caption{Module analysis on SemCo-VidOR and SemCo-AG datasets. 
	VAS, SAV, and SC denote Visual Augment Semantic, Semantic Augment Visual, and Sparse Coding, respectively.}
    \vspace{1mm}
    \scriptsize
	\resizebox{0.38\textwidth}{!}{
    \begin{tabular}{ccccccc}
		\midrule
		 \multicolumn{3}{c}{Modules} & \multicolumn{2}{c}{SemCo-VidOR} & \multicolumn{2}{c}{SemCo-AG} \\
		\cmidrule(lr){1-3} \cmidrule(lr){4-5} \cmidrule(lr){6-7}
		  VAS & SAV & SC & Acc & mAP & Acc & mAP \\
		\midrule
		 \checkmark &  &  & 30.08 & 28.29 & 67.33 & 76.36 \\
		   & \checkmark &  & 28.07 & 22.56 & 71.34 & 76.60 \\
		 \checkmark & \checkmark &  & 31.33 &27.79 & 74.73 & 77.92 \\
		 \checkmark & \checkmark & \checkmark & \textbf{36.84} & \textbf{33.59} & \textbf{75.50}& \textbf{80.94 }\\
		\midrule
	\end{tabular}%
    }
	\label{tab:module}%
\end{table}

\begin{table}[t]
    \centering
    \caption{Feature analysis on SemCo-AG and SemCo-VidOR datasets. 
    PVF, SVF, and OVF denote predicate, subject, and object visual features ($E^{p\prime}_{v}$, $E^{s}_{v}$, $E^{o}_{v}$), respectively.
    SeF and SpF denote semantic feature ($E^{\prime}_{se}$) and spatial feature ($E_{sp}$).}
    \vspace{1mm}
    \scriptsize
    \resizebox{0.48\textwidth}{!}{
    \begin{tabular}{ccccccccc}
        \midrule
        \multicolumn{5}{c}{Feature} & \multicolumn{2}{c}{SemCo-VidOR} & \multicolumn{2}{c}{SemCo-AG} \\
        \cmidrule(lr){1-5} \cmidrule(lr){6-7} \cmidrule(lr){8-9}
        PVF & SVF & OVF & SeF & SpF & Acc & mAP & Acc & mAP \\
        \midrule
        \checkmark &  &  &  &  & 30.96 & 28.77 & 71.32 & 72.35 \\
        \checkmark & \checkmark &  &  &  & 32.45 & 29.90 & 72.06 & 73.95 \\
        \checkmark &  & \checkmark &  &  & 31.88 & 29.15 & 73.89 & 74.65 \\
        \checkmark & \checkmark & \checkmark &  &  & 33.79 & 31.30 & 74.25 & 77.98 \\
        \checkmark & \checkmark & \checkmark & \checkmark &  & 35.85 & 33.04 & 73.96 & 77.10 \\
        \checkmark & \checkmark & \checkmark & \checkmark & \checkmark & \textbf{36.84} & \textbf{33.59} & \textbf{75.50} & \textbf{80.94} \\
        \midrule
    \end{tabular}
    }
    \label{tab:feature}
\end{table}

\subsubsection{Feature Analysis}
In this section, we analyze the impact of different components of visual relationship features $E_{r}$ given by Eq.~\ref{eq:concat} on predicting future predicates:
\begin{itemize}
    \item Predicate Visual Feature $E^{p \prime}_{v}$ (PVF) is used independently to predict future predicates. Corresponding to the first row in Table~\ref{tab:feature}.
    \item Subject Visual Feature $E^{s}_{v}$ (SVF) and the Object Visual Feature $E^{o}_{v}$ (OVF) are introduced separately for prediction. Corresponding to the second and third rows in Table~\ref{tab:feature}.
    \item The three visual features $E^{p \prime}_{v}$, $E^{s}_{v}$, $E^{o}_{v}$ are integrated into the complete visual feature for prediction. Corresponding to the fourth row in Table~\ref{tab:feature}.
    \item Semantic Feature $E_{se}$ (SeF) and the Spatial Feature $E_{sp}$ (SpF) are sequentially added to the visual feature, thereby verifying the importance of both. Corresponding to the fifth and sixth rows in Table~\ref{tab:feature}.
\end{itemize}

Results are presented in Table~\ref{tab:feature} indicate that the joint utilization of the five features significantly improves the Acc and mAP compared with the use of one or more features, for both SemCo-AG dataset and SemCo-VidOR dataset. 
In addition, the introduction of the subject visual feature leads to a greater improvement in SemCo-AG dataset, whereas the object visual feature plays a more significant role in SemCo-VidOR dataset. The possible reasons are that SemCo-VidOR dataset has more complex and diverse scenarios, involving a large number of relationships between animals, vehicles, and objects. In contrast, the SemCo-AG dataset mostly consists of daily activity scenarios, where interactive relationships are often determined by the actions and states of the subject. 
Moreover, the semantic feature is more important than the visual feature for SemCo-AG dataset. The possible reasons are that SemCo-AG dataset has fewer predicate categories, which will benefit the capture of the temporal dependencies by semantic feature. As a contract, the SemCo-VidOR dataset has more predicate categories, where the visual feature has a more effective representation ability. 
\begin{table}[t]
	\centering
	\caption{Parameter analysis in SemCo-VidOR dataset and SemCo-AG dataset. The parameters include: the size of local blocks $s$; the number of the sparse coding layers $L_{S}$; the number of the multi-head attention layers $L_{M}$.}
    \vspace{1mm}
    \scriptsize
        \resizebox{0.3\textwidth}{!}{
	\begin{tabular}{ccccc}
		\midrule
		\multirow{2}[4]{*}{} & \multicolumn{2}{c}{SemCo-VidOR} & \multicolumn{2}{c}{SemCo-AG} \\
		\cmidrule{2-5}          & Acc   & mAP   & Acc   & mAP \\
		\midrule
		$s$=2  & 36.07 & 30.48 & 75.92 & 76.96 \\
    $s$=3  & 34.13 & 29.19 & 73.95 & 79.20 \\
    $s$=4  & 36.84 & 33.59 & \textbf{75.50} & \textbf{80.94} \\
    $s$=5  & 37.07 & 31.36 & 74.04 & 79.86 \\
    $s$=6  & \textbf{37.14} & \textbf{33.60} & 73.92 & 80.01 \\
    \midrule
    $L_{S}$=1 & 34.73 & 31.91 & 74.24 & 79.32 \\
    $L_{S}$=2 & 36.84 & 33.59 & \textbf{75.50} & \textbf{80.94} \\
    $L_{S}$=3 & \textbf{37.29} & \textbf{34.06} & 74.86 & 80.43 \\
    \midrule
    $L_{M}$=1 & 35.01 & 32.85 & 73.13 & 76.43 \\
    $L_{M}$=2 & \textbf{36.84} & \textbf{33.59} & \textbf{75.50} & \textbf{80.94} \\
    $L_{M}$=3 & 36.12 & 33.53 & 72.36 & 79.90 \\
		\midrule
	\end{tabular}%
    }
	\label{tab:parameter}%
\end{table}%

\subsubsection{Parameter Analysis}
Important hyper-parameters are evaluated in this subsection. The parameters include: 1) the number of frames sparse $s$; 2) the number of the sparse coding layers $L_{S}$; 3) the number of the multi-head attention layers $L_{M}$. \par 
Results shown in Table \ref{tab:parameter} demonstrate that the best parameter values vary across different datasets. {Note that the default configurations of parameters are ($s=4$, $L_{S}=2$, $L_{M}=2$) for SemCo-VidOR dataset and ($s=4$, $L_{S}=2$, $L_{M}=2$) for SemCo-AG dataset. In the parameter analysis, except for the testing parameter, other parameters are set according to the default settings.} 
As for the size of lock blocks $s$, the model reaches the highest Acc value when $s=4$ on SemCo-AG dataset. However, for SemCo-VidOR dataset, models with a larger $s$ value $s=6$ are obviously more effective. The number of sparse coding layers $L_{S}$ is similar, SemCo-AG dataset achieves the best results when $L_{S}=2$, and SemCo-Vidor dataset achieves the best results when $L_{S}=3$. The possible explanation is that the videos in SemCo-Vidor dataset contain richer objects, and increasing the number and density of sparse coding is beneficial to capture complex spatial-temporal interactions. 
On both datasets, the model with $L_{M}=2$ reaches the highest performance. When $L_{M}=1$, the Acc drops 1.83\% and mAP drops 0.74\% for SemCo-VidOR dataset, while the Acc drops 2.37\% and mAP drops 4.51\% for SemCo-AG dataset. This suggests that setting $L_{M}=2$ provides a sufficient depth for capturing temporal dependencies without introducing excessive complexity or overfitting.

\section{CONCLUSION}
\label{sec:concluding}

In this paper, we propose a benchmark specifically designed for semantic coherent visual relationship forecasting, named SemCoBench. Two datasets with closely-related, spatio-temporally localized visual relation annotations, SemCo-AG and SemCo-VidOR, are established to advance research in visual relationship forecasting and video understanding. Moreover, we introduce a baseline method named the Semantic Coherent Transformer (SemCoFormer), which leverages cross-modal cues to extract coherent features and applies sparse coding to facilitate the model's focus on cross-frame relationship dynamics. Experimental results on SemCoBench demonstrate that capturing semantic coherence plays a crucial role in achieving reasonable, fine-grained, and diverse visual relationship understanding.

 \bibliographystyle{elsarticle-num-names} 
 \bibliography{main}

\end{document}